\RequirePackage{fix-cm}
\documentclass[twocolumn,natbib]{svjour3}

\usepackage{amssymb, amsfonts, amsthm, amsmath, amsbsy, latexsym,bbm,url}
\usepackage[dvipsnames]{xcolor}
\usepackage[pdftex]{graphicx}
\usepackage{mathrsfs}
\usepackage[pdfborder={0 0 0}]{hyperref}
\usepackage{multirow}
\usepackage{placeins}
\smartqed

\theoremstyle{plain}
\newtheorem{thm}{Theorem}[section]

\theoremstyle{definition}
\newtheorem{defn}[thm]{Definition}

\newcommand{\bC}{\mathbb{C}}

\newcommand{\bR}{\mathbb{R}}

\newcommand{\bZ}{\mathbb{Z}}

\newcommand{\ip}[2]{\langle#1,#2\rangle}

\newcommand{\abs}[1]{\left\vert#1\right\vert}
\newcommand{\absip}[2]{\left\vert\langle#1,#2\rangle\right\vert}

\begin{document}
\sloppy
\title{Shearlet-Based Detection of Flame Fronts}

	\author{Rafael Reisenhofer \and Johannes Kiefer \and Emily J. King}

	\institute{Rafael Reisenhofer \at Computational Data Analysis, Fachbereich 3, Universit\"at Bremen, Postfach 330440, 28334 Bremen
		\and
		Johannes Kiefer \at
		Technische Thermodynamik, Fachbereich 4, Universit\"at Bremen, Badgasteiner Str. 1, 28359 Bremen \\
		Tel.: +49-421-218-64777\\
		Fax: +49-421-218-64771\\
		\email{jkiefer@uni-bremen.de}
		\and
		Emily J. King \at
		Computational Data Analysis, Fachbereich 3, Universit\"at Bremen, Postfach 330440, 28334 Bremen
	}
		
	\maketitle

\begin{abstract}Identifying and characterizing flame fronts is the most common task in the computer-assisted analysis of data obtained from imaging techniques such as planar laser-induced fluorescence (PLIF), laser Rayleigh scattering (LRS), or particle imaging velocimetry (PIV). We present Complex Shearlet-Based Ridge and Edge Measure (CoShREM), a novel edge and ridge (line) detection algorithm based on complex-valued wavelet-like analyzing functions -- so-called complex shearlets -- displaying several traits useful for the extraction of flame fronts.  In addition to providing a unified approach to the detection of edges and ridges, our method inherently yields estimates of local tangent orientations and local curvatures. To examine the applicability for high-frequency recordings of combustion processes, the algorithm is applied to mock images distorted with varying degrees of noise and real-world PLIF images of both OH and CH radicals. Furthermore, we compare the performance of the newly proposed complex shearlet-based measure to well-established edge and ridge detection techniques such as the Canny edge detector, another shearlet-based edge detector, and the phase congruency measure.
\end{abstract}

\section{Introduction}
The development of laser combustion diagnostics employing planar imaging techniques in the 1970s and 80s has transformed combustion research \citep{dye1982,fou1986,ald2011,thu2013}. Using a light sheet to illuminate an entire two-dimensional cross-section of a flame and imaging the laser-induced emission onto a camera provides spatially correlated information in contrast to pointwise scanning. In particular, the use of short-pulse laser sources and gated cameras enables imaging on time scales that are shorter than flow and diffusion phenomena, and hence a true snapshot of a flame can be taken. Consequently, studying transient phenomena is possible by capturing flame structures under turbulent conditions. However, processing and evaluating such images is a challenge. Appropriate methods must be reproducible, accurate, and quantitative. In addition, the information desired should be available within a reasonable period of time. This is particularly important when large data sets need to be processed.

\

The most common task is to identify and to characterize the flame front in an image recorded by planar laser-induced fluorescence (PLIF) \citep{swe2009}, laser Rayleigh scattering (LRS) \citep{pfa2007} or particle imaging velocimetry (PIV) \citep{pfa2007}. Regarding data processing, this task comes down to an edge detection problem. Needless to say, the edge detection step is crucial, as a slightly differently detected edge may suggest significantly different flame parameters, e.g. in terms of the flame front curvature.

\

The majority of existing approaches for detecting the flame front in an image are based either on direct binarization \citep{kie2008,haq2002} or on local intensity gradients \citep{sla2015,bay2012}. When direct binarization is applied, an intensity threshold filter is used delivering a binary image containing areas of zeroes and ones, representing unburnt and burnt regions. The boundary between the two is the flame front, from which further information can be derived. In gradient-based methods, the first step is to convert the initial image by computing an approximation of the gradient pixel-by-pixel. Then a local or global threshold is applied in order to discriminate between the steep gradients typical for the flame front and less pronounced structures. This may be a simple threshold or a sophisticated combination of multiple thresholds via hysteresis \citep{Can1986}. Subsequently, the remaining flame front data points can be fitted with a mathematical function, from which parameters such as curvature and flame surface density can be derived eventually. Pre-processing the original images with filters for noise reduction and contour enhancement may be required in order to improve the clearness and robustness of the flame front detection \citep{sla2015,swe2009,mal2000}. However, when the signal-to-noise ratio is low or the edge to be detected is not sufficiently steep, the common data processing algorithms may reach their limits. Further, the typical results must be further processed (e.g., by fitting a cubic spline to the detected edge) in order to obtain geometric information like curvature \citep{pfa2007}. Finally, traditional edge-detection algorithms are not capable of detecting ridges (lines) as coherent structures, which is problematic when analyzing images of short-lived radicals like {CH} and {HCO}.  Instead, completely different ridge-detection methods have to be applied, which often are based on approximating local optima  \citep{Lin98,SANVG} or matching ridges to set shapes like circles \citep{DuHa72}.

\

To overcome the limitations of existing flame image analysis tools, we propose the application of an algorithm which we have named Complex Shearlet-Based Ridge and Edge Measure (CoShREM) for the detection and analysis of edges and ridges, based on a so-called shearlet transformation. Shearlets were introduced a decade ago \citep{KLLW2005} to handle geometric structures in 2D data, and a novel method to use so-called complex shearlets in edge detection initially appeared in  \citep{Rei14} and was fine-tuned in \citep{KRKLLH}.  A description of the basic mathematical intuition behind CoShREM will be given in Section~\ref{sec:math}. This is the first paper, to the best of the authors' knowledge, to use any sort of shearlet-based method (complex-shearlet-based or otherwise) in the processing of flame images.

\

In the present paper, we investigate the potential of shearlet transformations for evaluating data from planar laser diagnostics. As a first step, in Sections~\ref{sec:mock_edges} and \ref{sec:mock_lines} mock images with clearly defined structures, which have been corrupted by blurring, Gaussian noise and Poisson noise of varying levels, are processed in order to allow a systematic assessment of the method. The mock images were generated such that they represent the characteristics of typical flame data. One set of images exhibits thin ridges, which are commonly observed when short-lived radicals such as CH and HCO are visualized using PLIF. The other data set shows broader areas (edge detection), which are characteristic of LRS and PIV data, as well as PLIF images of long-lived radicals like OH. The new shearlet-based method will be shown to work well in detecting edges \emph{and} ridges over a range of noise levels and amounts of added blur, in addition to giving geometric information -- in particular tangent slope and curvature -- about the edges.  However, we shall also show that, given the correct parameters, the classical method of \citet{Can1986} is also quite successful at detecting edges (but not ridges or curvature). Thus, in addition to introducing a novel method of edge and ridge detection, a discussion of how to properly use a very classical method will be presented.  In the second step, Section~\ref{sec:real}, CoShREM is applied to {CH} and {OH} PLIF images of a turbulent jet flame in order to demonstrate its performance in the analysis of experimental data.

\

The main contributions of this paper are the introduction of a new method of flame image analysis and a systematic analysis of methods of edge detection, ridge detection, and local curvature calculation.  Through this analysis, the new shearlet-based method CoShREM is shown to be robust to noise and blurring when detecting edges or ridges and also determining curvature.  The proper way to parameterize the Canny method to obtain good results for basic edge detection is also presented.

\section{Mathematical Background}\label{sec:math}
A very common approach to edge detection is to say that edges occur where the gradient -- a generalization of the derivative to higher dimensions -- is high; so one approximates the gradient and looks for where it is large \citep{Rob63,Prew70,SF68,PED90}.   Simply approximating the gradient is very sensitive to noise, so the image is typically smoothed with a Gaussian kernel \citep{Can1986} or is approximated in a multiscale manner using wavelets \citep{MaHw92,MaZh92} or even shearlets \citep{YLEK2009}. The method presented in this paper is different in that it does not attempt to approximate the gradient.  In order to explain the mathematical intuition of the approach, we begin by defining some notation.

\

We shall denote the set of integers by $\bZ$, the set of real numbers by $\bR$, the set of complex numbers by $\bC$, and the square root of $-1$ by $i$.  For functions of finite energy $f$ and $g$, $\ip{f}{g}$ represents the standard inner product, which is conjugate linear in the second component
\[
\ip{f}{g} = \int_{-\infty}^{\infty} f(x) \overline{g(x)} dx.
\]

\

\begin{figure}[h]
\centering
\begin{tabular}{cccc}
\includegraphics[width=0.22\linewidth]{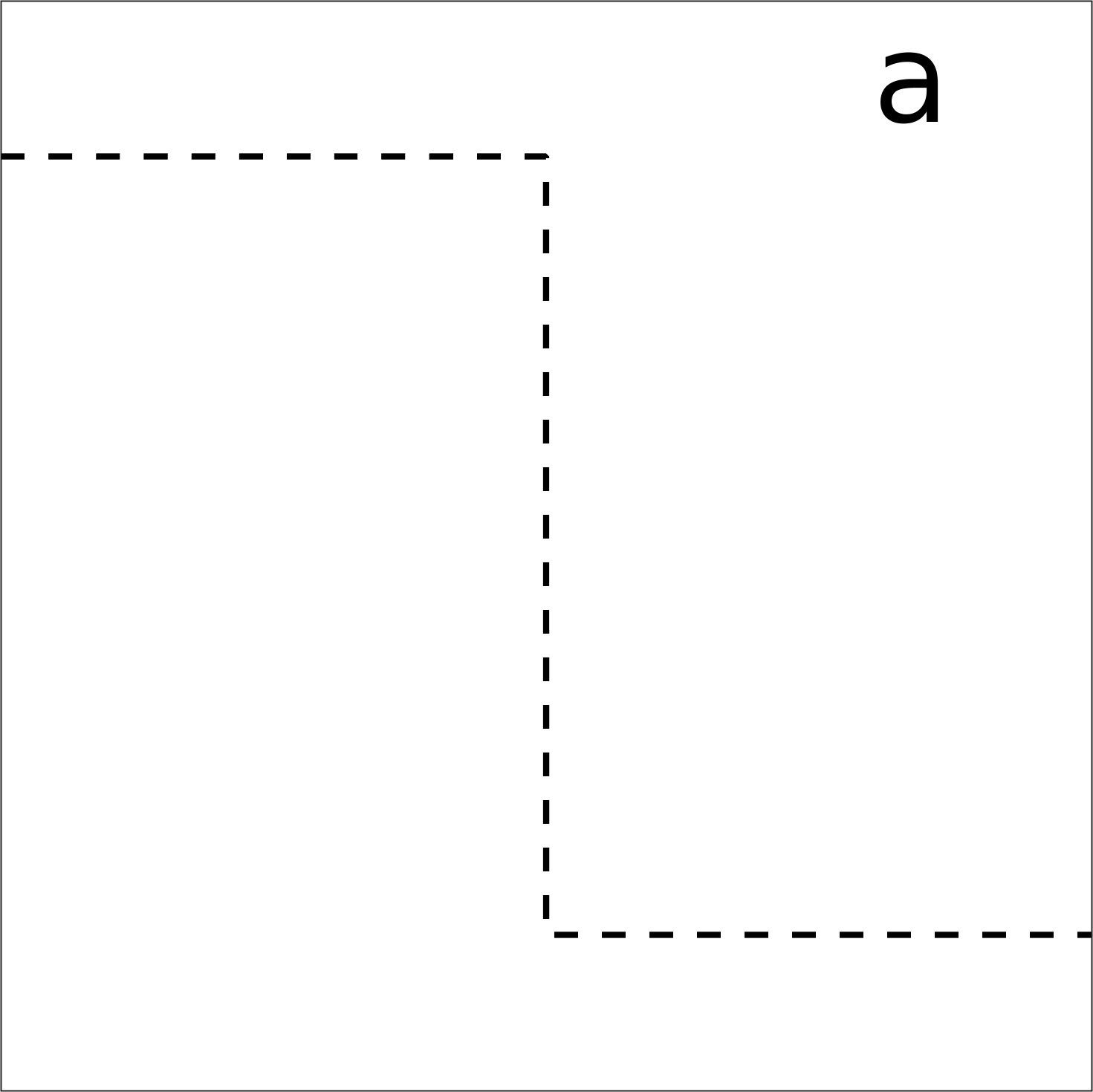}&
\includegraphics[width=0.22\linewidth]{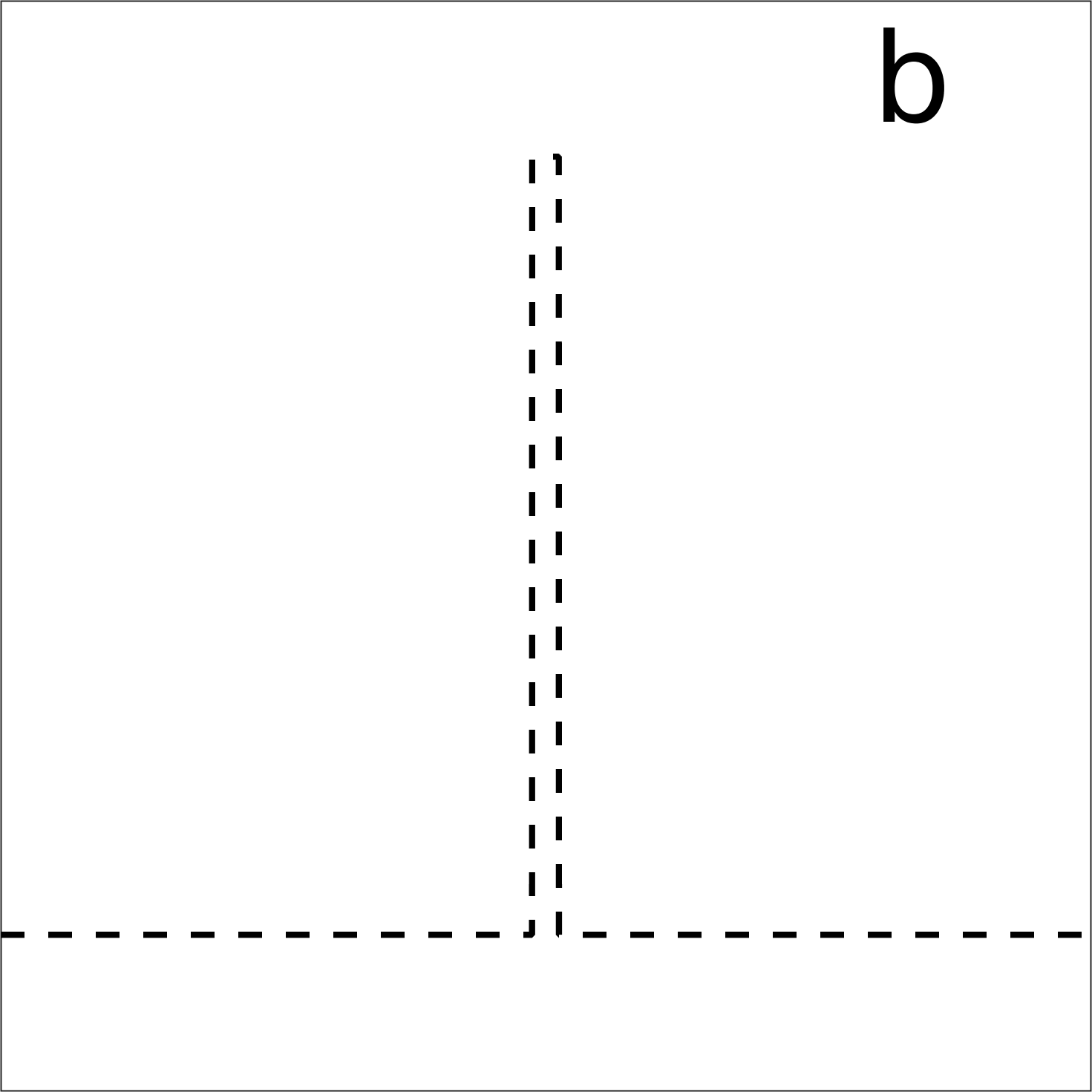}&
\includegraphics[width=0.22\linewidth]{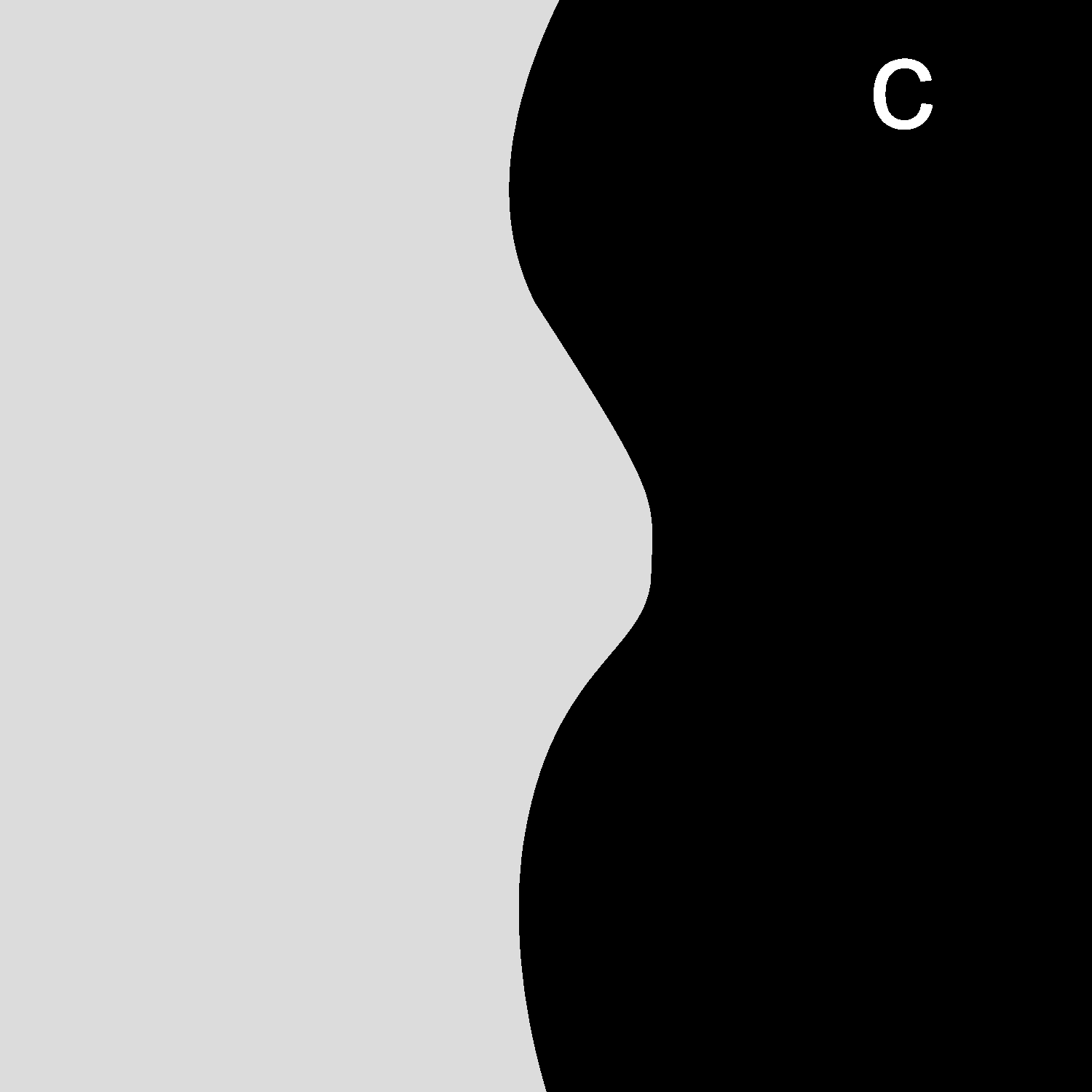}&
\includegraphics[width=0.22\linewidth]{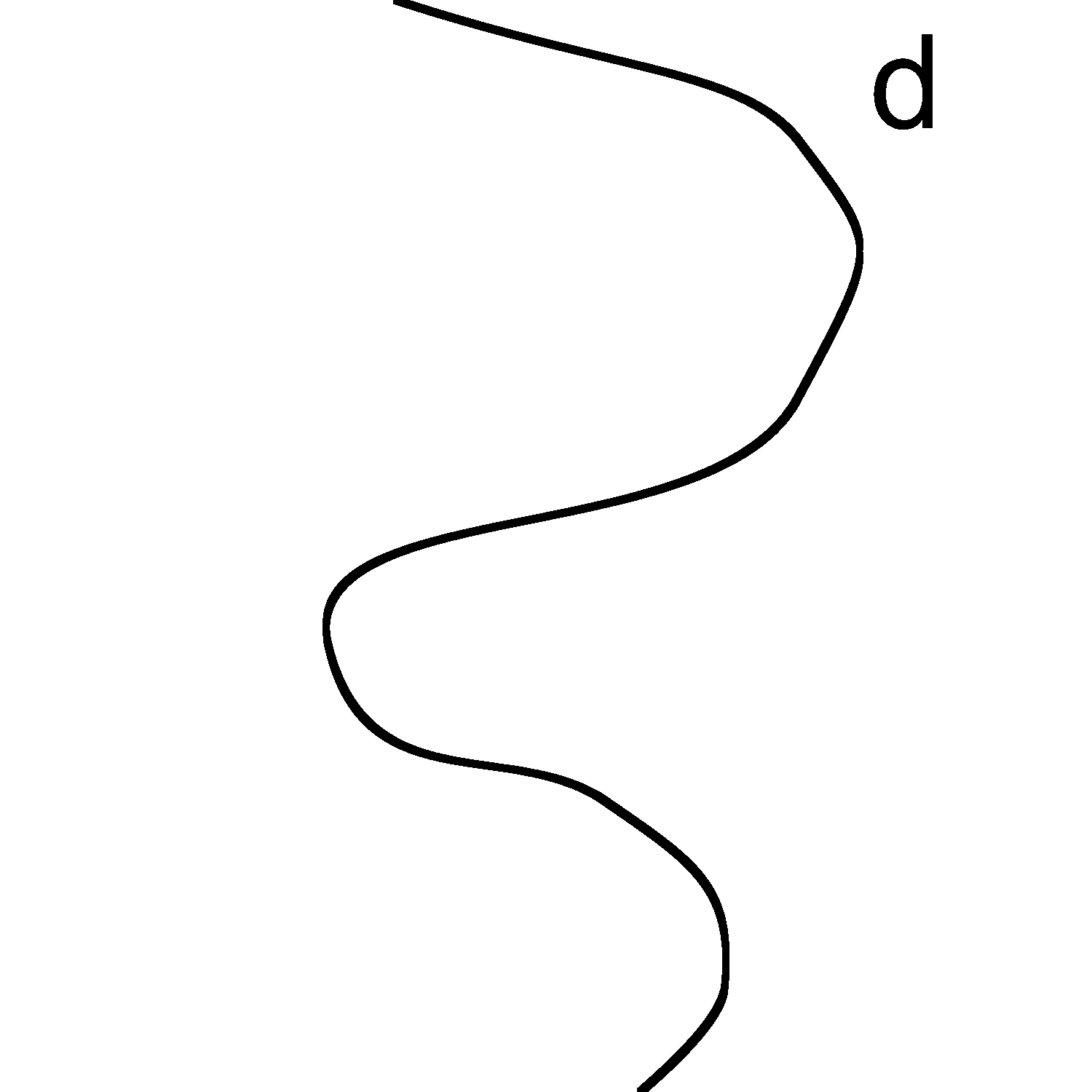}
\end{tabular}
\caption{(a) A 1D model edge (jump). (b) A 1D model ridge. (c) A 2D model edge.  (d) A 2D model ridge.}
\label{fig:edgeridge}
\end{figure}

We wish to detect both edges and ridges.  Examples of what is meant by ``edge'' and ``ridge'' in one and two dimensions may be found in Figure~\ref{fig:edgeridge}.  Figure~\ref{fig:edgeridge}(a) is a 1D model edge, (b) a 1D model ridge, (c) a 2D model edge, and (d) a 2D model ridge.
\setlength{\tabcolsep}{3pt}
\setlength{\tabcolsep}{6pt}
The approach we shall use involves two key ideas: phase congruency and shearlets. First, the phase congruency approach \citep{MoOw1987,MRBO1986,Kov2000,Kov1999} to edge and ridge detection is strengthened (see also \citep{DeDu00}).  Then, it is implemented using systems of complex shearlets \citep{KLLW2005,Sto2013,Rei14}, which are wavelet-like functions that yield geometric information.  A detailed explanation of the mathematics behind the method will be in a forthcoming paper with a shorter synopsis in \citep{KRKLLH} and an earlier version appearing in \citep{Rei14}.

\

We first summarize the intuition behind phase congruency, which is a type of harmonic-analysis-based signal processing. That is, given a signal or image represented as a function $f$, we want to obtain information about $f$ by considering the inner products of $f$ with some collection of functions $\{\varphi_j\}$, namely, the set $\{\ip{f}{\varphi_j}\}$.  Examples of common functions $\varphi_j$ used in this context are complex exponential functions (used in Fourier analysis) and wavelets.  For simplicity's sake, we describe the approach in one dimension.  Further, while the original papers about phase congruency \citep{MoOw1987,MRBO1986} dealt with Fourier-based phase congruency, this is highly sensitive to noise and the spacing between edges.  Thus, for brevity, we shall explain wavelet-based phase congruency, which avoids these issues.

\

\begin{figure}[h]
\centering
\begin{tabular}{cc}
\includegraphics[width = 0.45\linewidth]{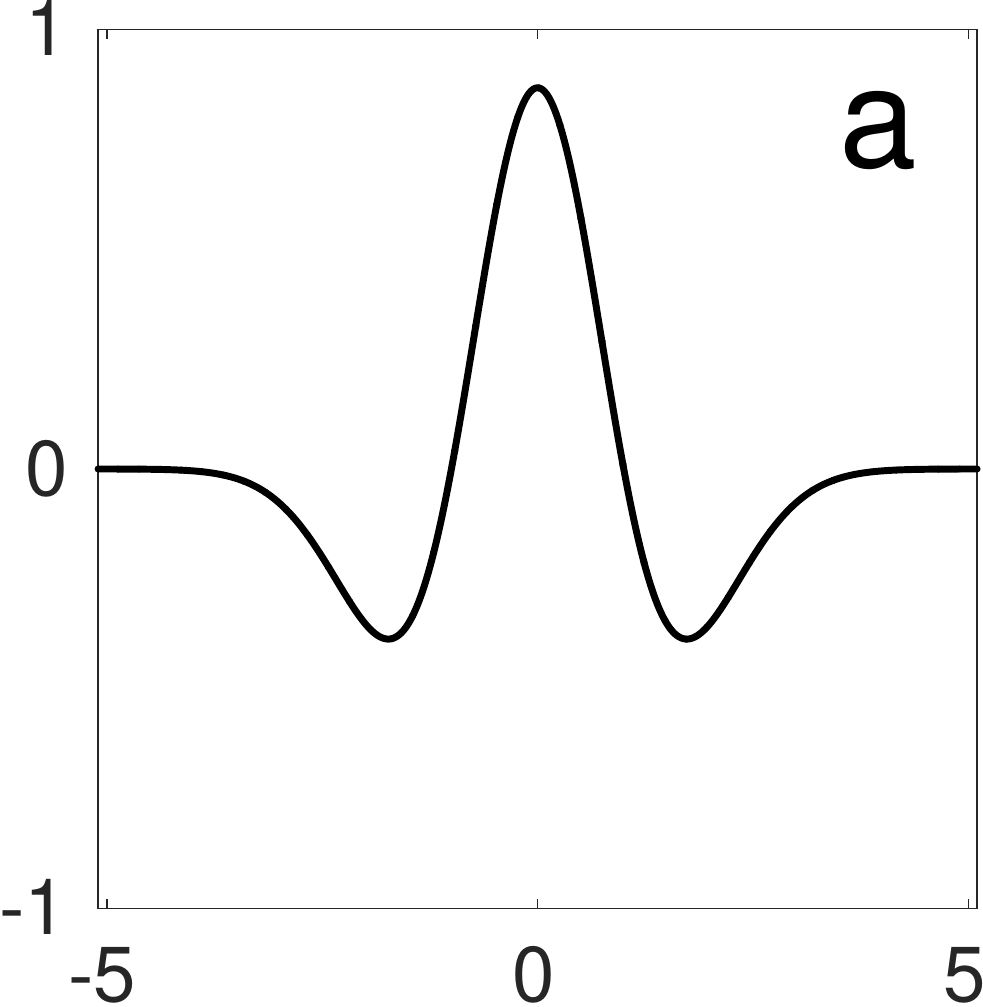}&\includegraphics[width = 0.45\linewidth]{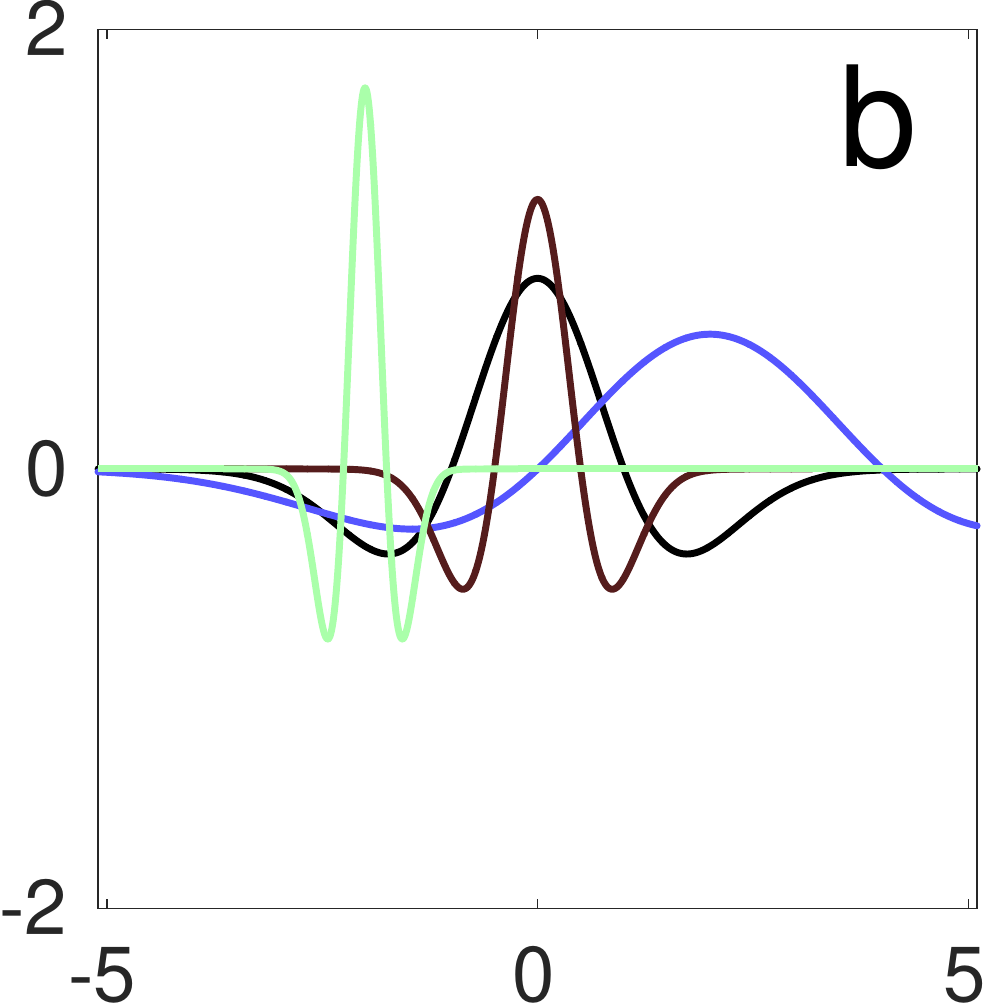}\\
\includegraphics[width = 0.45\linewidth]{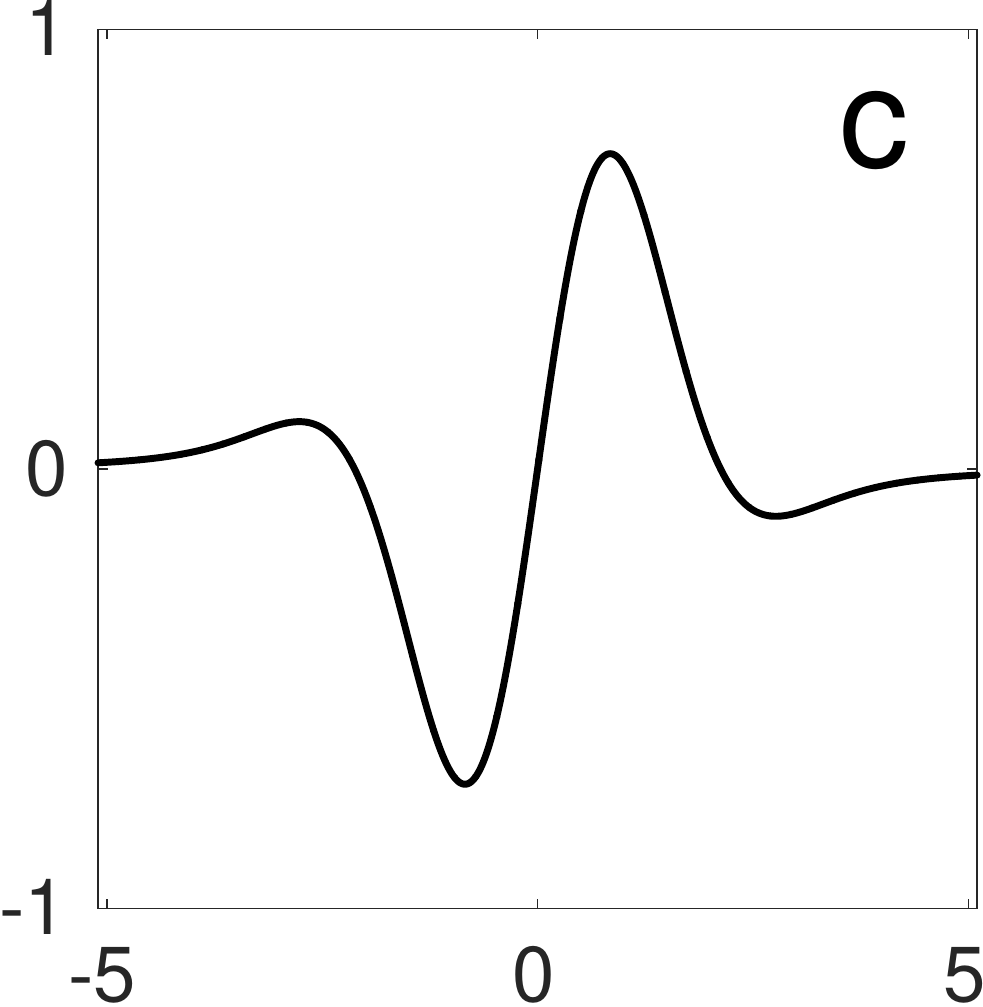}&\includegraphics[width = 0.45\linewidth]{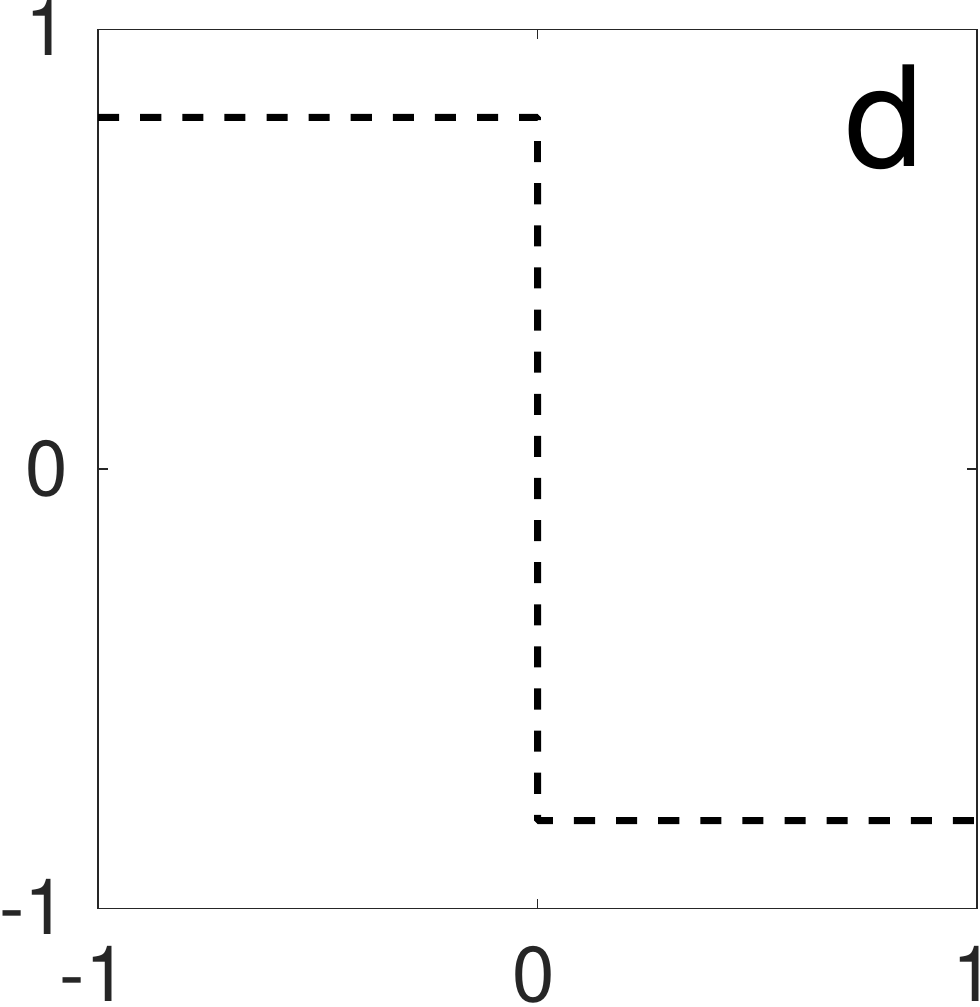}
\end{tabular}
\caption{(a) The so-called Mexican hat wavelet, an even-symmetric real-valued wavelet $\psi^{(e)}$. (b) Various shifts and dilations of the Mexican hat wavelet. $\psi^{(e)} = \psi_{1,0}^{(e)}$ is in black, $\psi_{1/2,0}^{(e)}$ in dark red, $ \psi_{2,2}^{(e)}$ in blue, and $ \psi_{1/4,-2}^{(e)}$ in light green. (c) The odd-symmetric wavelet $\psi^{(o)}$ paired with the Mexican hat wavelet.  (d) An idealized edge in one dimension.}
\label{fig:evenodd}
\end{figure}

Wavelets were first introduced by Haar \citep{Haar09,Haar10} in the early 20th century and rose to prominence about 80 years later. Unlike the Fourier basis, formed from periodic sines and cosines, wavelet collections are built from functions which take values close to zero away from the origin and which are stretched and shrunk.  This means that inner products of a function with sines and cosines yields information about periodic phenomenon, while inner products with wavelets result in a characterization of local traits. See, for example \citep{Dau92} for a general reference.  Given a function $\psi: \bR \rightarrow \bC$ which satisfies certain mathematical properties, we define a wavelet system as
\begin{equation}\label{eqn:1dwave}
\{ \psi_{a,y}(x) := (\sqrt{a})\psi( a(x - y)): a > 0, y \in \bR\}.
\end{equation}
That is, we form the system by dilating by $a$ and translating by $y$ the function $\psi$. Figure~\ref{fig:evenodd}(a) contains an example of a wavelet, and Figure~\ref{fig:evenodd}(b) shows various shifts and translates of that wavelet.  The quantities of interest for us are the so-called \emph{wavelet coefficients}, 
\[
\ip{f}{\psi_{a,y}}, \quad a > 0, y \in \bR.
\]
Often in applications instead of dilating by all possible $a > 0$ and translating by all possible $y \in \bR$, one dilates by $2^n$ for $n \in \bZ$ and shifts by $k \in \bZ$; however, for our purposes, we gain more information by allowing more dilations and translations. Furthermore, since we will always use $\psi$ which are centered at the origin, the parameter $y$ tells one where -- regardless of the value of $a$ -- $\psi_{a,y}$ is centered, namely, at $y$.

\

\begin{figure}[h]
\centering
\begin{tabular}{cc}
\includegraphics[height = 0.4\linewidth]{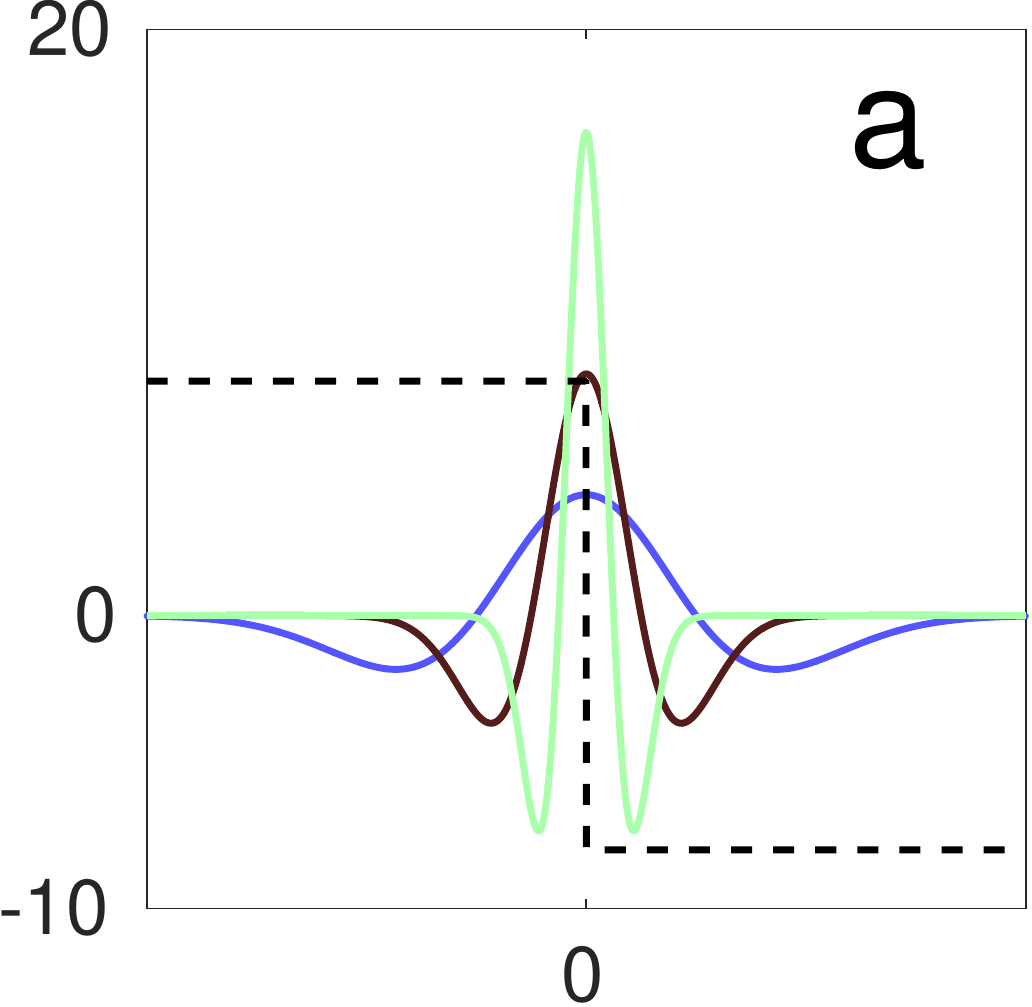}&\includegraphics[height = 0.4\linewidth]{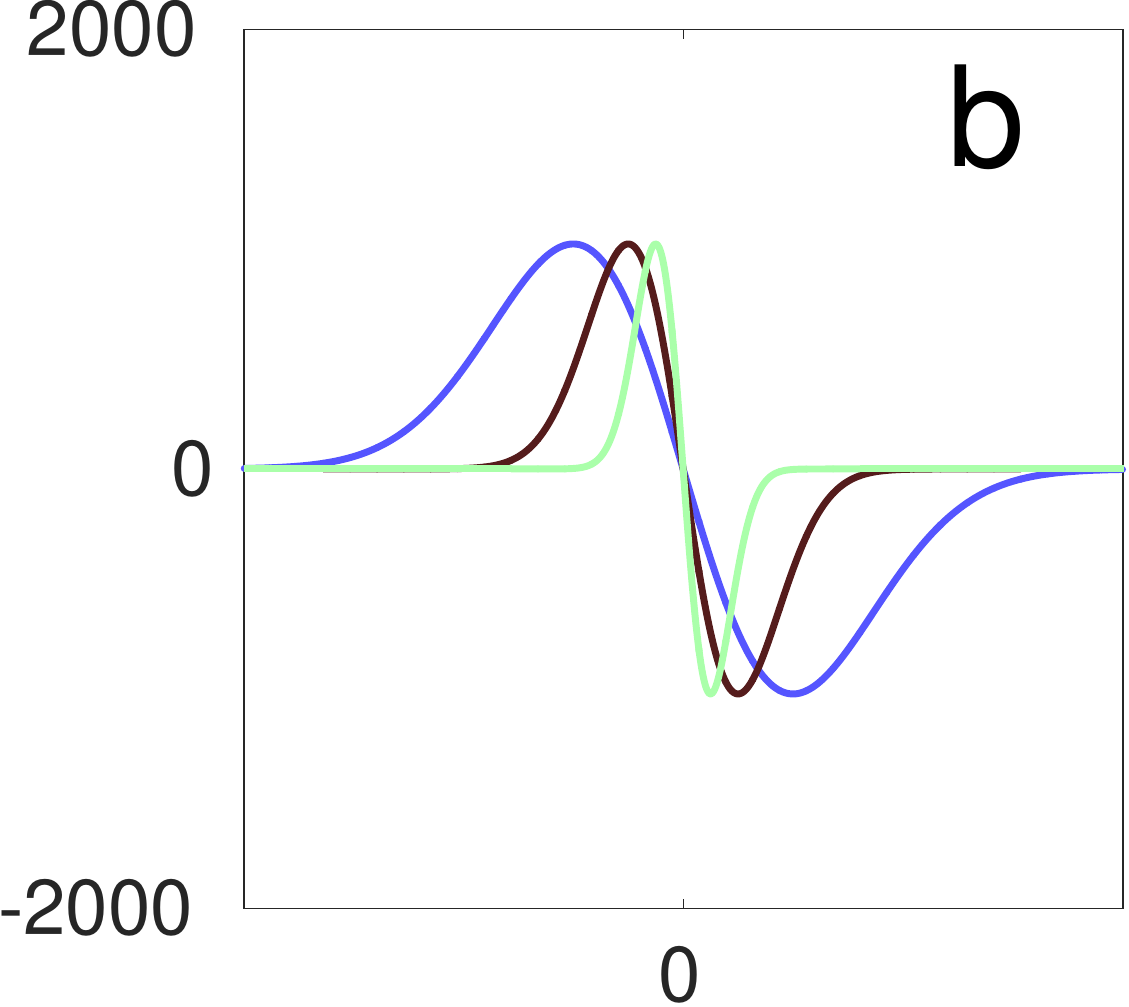}\\
\includegraphics[height = 0.4\linewidth]{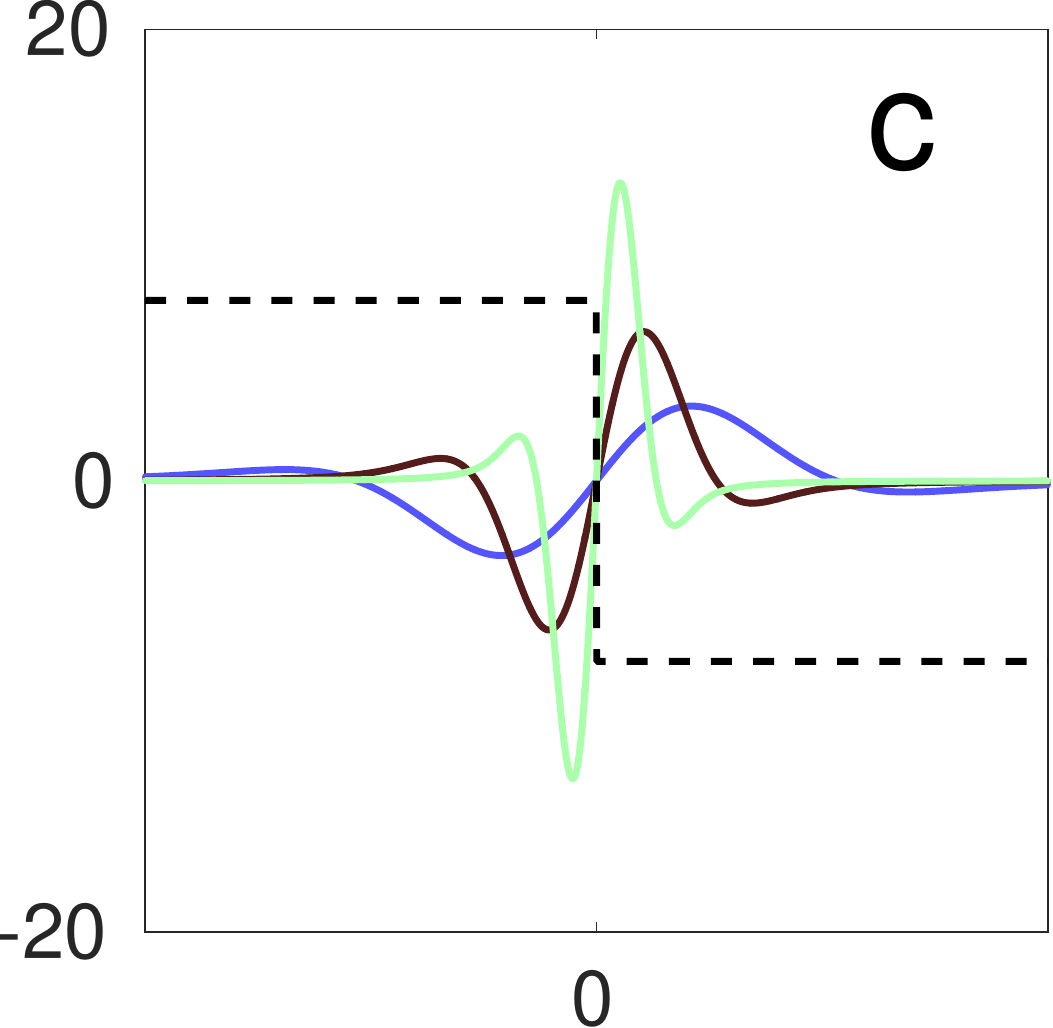}&\includegraphics[height = 0.4\linewidth]{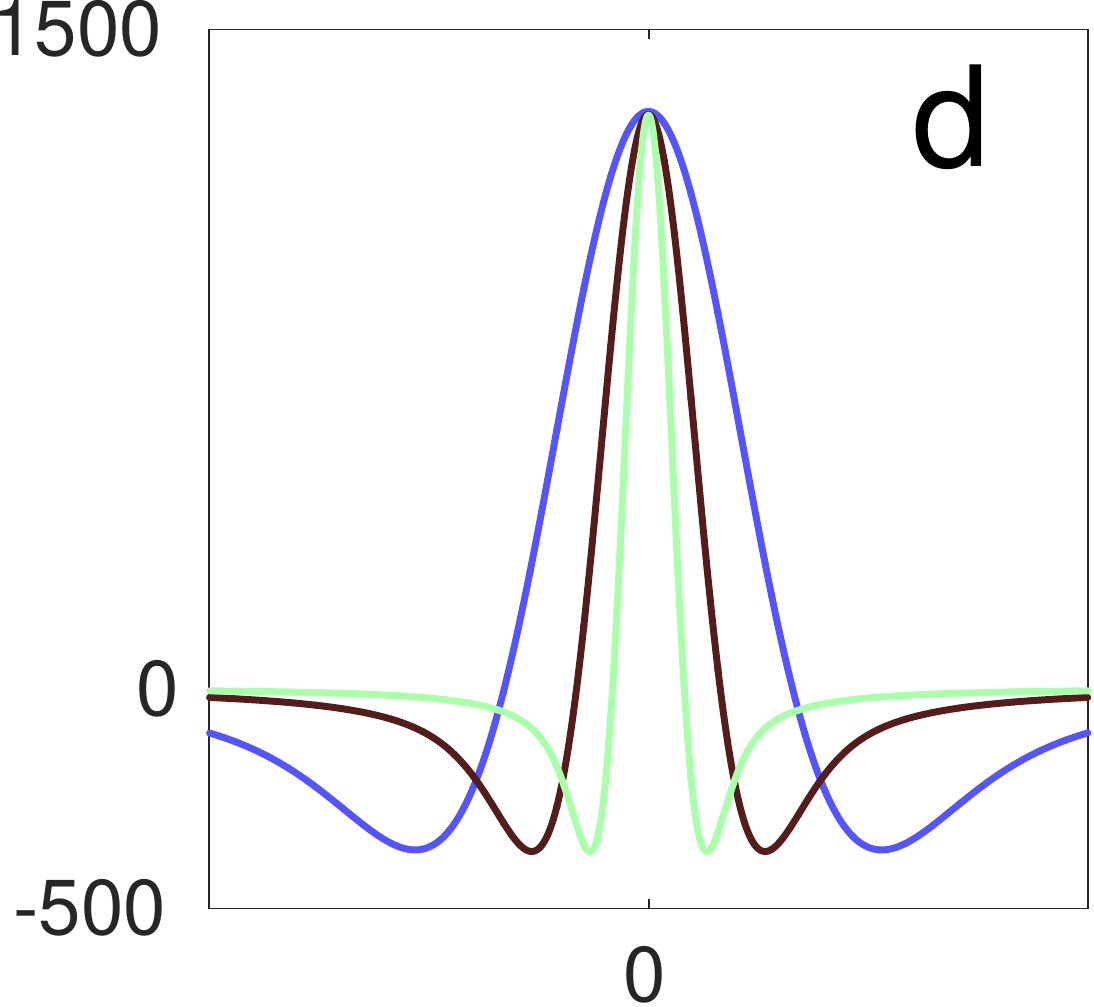}
\end{tabular}
\caption{(a) $\psi_{a,0}^{(e)}$, dilations of the even-symmetric wavelet centered on the edge. (b) A graph of the wavelet coefficients of the differently dilated even-symmetric wavelets at different positions.  The horizontal axis is the shift $y$ and the vertical axis is the value of $\ip{f}{\psi_{a,y}^{(e)}}$. (c) $\psi_{a,0}^{(o)}$, dilations of the odd-symmetric wavelet centered on the edge. (d) A graph of the wavelet coefficients of the differently dilated odd-symmetric wavelets at different positions.  The horizontal axis is the shift $y$ and the vertical axis is the value of $\ip{f}{\psi_{a,y}^{(o)}}$.}
\label{fig:coeffs}
\end{figure}

We shall use a special type of $\psi$ called a complex wavelet. A complex wavelet is formed as
\[
\psi^{(c)} = \psi^{(e)} + i \psi^{(o)},
\]
where $\psi^{(e)}$ and $ \psi^{(o)}$ are both real-valued, $ \psi^{(e)}$ is even-symmetric (that is, $ \psi^{(e)}(-x) =  \psi^{(e)}(x)$ for all $x \in \bR$), $ \psi^{(o)}$ is odd-symmetric (that is, $ \psi^{(o)}(-x) =  -\psi^{(o)}(x)$ for all $x \in \bR$), and $\psi^{(e)}$ and $ \psi^{(o)}$ have a particular relationship to each other \citep{Kin1999,Sel01,SeAb2004}. In Figure~\ref{fig:evenodd}, one can see examples of $\psi^{(e)}$ (a) and $ \psi^{(o)}$ (c), as well as the model for a one-dimensional edge (d) that we shall use as a motivating example.  
\renewcommand{\arraystretch}{3}
We shall call $\ip{f}{\psi_{a,y}^{(e)}}$ the \emph{even-symmetric coefficients} and $\ip{f}{\psi_{a,y}^{(o)}}$ the \emph{odd-symmetric coefficients}. The underlying idea of wavelet-based phase congruency is that when the wavelets $\psi^{(e)}_{a,y}$ and $\psi^{(o)}_{a,y}$ are properly normalized and there is an idealized edge in a function $f$ at position $y$, $\ip{f}{\psi_{a,y}^{(e)}} = 0$ for all $a > 0$ and $\ip{f}{\psi_{a,y}^{(o)}}$ is constant and non-zero for all $a> 0$. This can be seen in Figure~\ref{fig:coeffs}.
\renewcommand{\arraystretch}{1}
Figure~\ref{fig:coeffs}(a) shows three dilations of $\psi^{(e)}$ centered on the edge, while Figure~\ref{fig:coeffs}(b) shows the values of the inner product of $f$ with shifted versions of the three dilations of $\psi^{(e)}$.  In particular, notice that when each of the dilated wavelets is centered on the edge, their inner product with $f$ is zero.  As the dilated wavelets are shifted away from the edge, the inner product increases or decreases rapidly away from zero.  On the other hand, if we consider dilated and shifted odd-symmetric wavelets as in Figure~\ref{fig:coeffs}(c), we can see in Figure~\ref{fig:coeffs}(d) that their inner product with $f$ reaches a maximum, in fact the same maximum, when they are centered on the edge.  As they are shifted away from the edge, the value of the inner product drops rapidly from that maximum.  Of course, in real data, edges are not usually so sharp and are corrupted by noise.  However, it is still true that for $a>0$, $\ip{f}{\psi_{a,y}^{(e)}}$ is close to zero if $f$ has an edge at $y$ and $\ip{f}{\psi_{a,y}^{(o)}}$ is both ``large'' relative to nearby values of $y$ and approximately equal across different values of $a$.  We use these three properties to define a measure between $0$ and $1$ of -- in some sense -- the likelihood that a pixel is part of an edge.
\begin{defn}
Given a pair of even-symmetric and odd-symmetric wavelets $\psi^{(e)}$ and $\psi^{(o)}$ which satisfy certain hypotheses (in particular, a normalized version of the wavelets in Figure~\ref{fig:evenodd}(a) and (c) work), we choose a set of $J$ positive dilations $\{a_j\}_{j \in \{1, 2, \hdots, J\}}$ and a very small $\epsilon > 0$. Then we define for a 1D signal $f: \bR \rightarrow \bC$
\[
\tilde{E}(y) = \frac{\abs{\sum_{j=1}^J \ip{f}{\psi_{a_j,y}^{(o)}}}-\sum_{j=1}^J \absip{f}{\psi_{a_j,y}^{(e)}}}{J \cdot \max_{j \in \{1,2,\hdots,J\}} \absip{f}{\psi_{a_j,y}^{(o)}} +\epsilon}
\]
and
\begin{equation}\label{eqn:nozero}
E(y) = \max\{\tilde{E}(y),0\}.
\end{equation}
\end{defn}
$E$ is a function of $y$, the location, which yields a value between $0$ and $1$ showing likelihood that there is an edge at $y$.  Considering an idealized edge as in Figure~\ref{fig:evenodd}(d), we can see from Figure~\ref{fig:coeffs}(b) and (d) that when $y$ is the location of the jump, regardless of which $a_j$ are chosen, $\ip{f}{\psi_{a_j,y}^{(e)}}$ are all $0$ and each $\ip{f}{\psi_{a_j,y}^{(o)}}$ is equal, hence
\[
\frac{\sum_j \ip{f}{\psi_{a_j,y}^{(o)}}-\sum_j \ip{f}{\psi_{a_j,y}^{(e)}}}{J \cdot \max_{j } \ip{f}{\psi_{a_j,y}^{(o)}}} = \frac{{\sum_j \ip{f}{\psi_{a_j,y}^{(o)}}}}{J \cdot \max_{j} \ip{f}{\psi_{a_j,y}^{(o)}}} = 1.
\]
The absolute values are important because the edge may be a ``negative edge''; that is, the edge to be detected looks like the mirror image of  Figure~\ref{fig:evenodd}(d). The epsilon is added to prevent division by zero. Note that when $y$ is on an edge, the odd-symmetric coefficients are large and equal and the even-symmetric coefficients are zero. So as long as $\epsilon$ is small, $E(y)$ is still basically $1$.  If the even-symmetric coefficients are larger than the odd-symmetric coefficients, resulting in a negative $\tilde{E}(y)$, then there is essentially no chance that there is an edge located at $y$.  Thus we simply set the value to be $0$ as in Equation~\ref{eqn:nozero}.

\

\begin{figure}[h]
	\centering
	\begin{tabular}{ccc}		
		\includegraphics[width=0.3\linewidth]{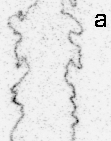}&
		\includegraphics[width=0.3\linewidth]{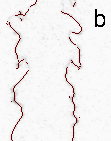}&
		\includegraphics[width=0.3\linewidth]{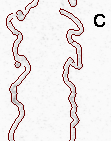}
	\end{tabular}
	\caption{(a) A small portion of the image processed in Figure~\ref{fig:real_lines}. (b) Ridges detected by CoShREM. (c) Results of the Canny edge detector. The figure illustrates that gradient-based methods such as the Canny edge detector can only detect the boundary curves of ridges but fail to recognize ridges as coherent structures.}
	\label{fig:ridgevsedgemeasure}
\end{figure}

In order to get a ridge measure, one notes that when the wavelets are centered on a ridge (Figure~\ref{fig:edgeridge}(b)) at $y$, across scales $a>0$, each $\ip{f}{\psi_{a,y}^{(e)}}$ is large, while  $\ip{f}{\psi_{a,y}^{(o)}}$ is close to zero. So basically, one more-or-less switches the roles of $\psi_{a,y}^{(e)}$ and $\psi_{a,y}^{(o)}$ in the above definition to get a ridge measure. This trick leads to a measure that truly picks up the structure of ridges, which is something gradient-based edge detectors by construction fail to do. Gradient-based detectors like Canny can only pick up the boundary edges of a ridge rather than the ridge itself, as seen in Figure~\ref{fig:ridgevsedgemeasure}.

\

This edge measure works quite well in one dimension, but we are of course interested in finding the edges of two-dimensional images.  We want to generate a system by manipulating a function $\psi :\bR^2 \rightarrow \bC$ that is wavelet-like.  While it is possible to consider the most obvious generalization of Equation~\ref{eqn:1dwave} to two dimensions
\begin{equation}\label{eqn:2dwave}
\{ \psi_{a,y}(x) := \psi( a(x - y)): a > 0, y \in \bR^2\},
\end{equation}
such systems do not handle curvilinear structures well due to their isotropic nature.  That is to say, when dilating by $a$ to get $\psi_{a,y}$, the function is stretched the same in each direction, so angled edges cannot be picked up very well.  A number of systems like curvelets \citep{CaDo04}, ridgelets \citep{CG02}, contourlets \citep{DoV03},  bandlets \citep{LePM}, wedgelets \citep{Don99}, and shearlets \citep{GKL06,KL09,KLLW2005} have been created over the last decade or so in an attempt to construct wavelet-like systems that give more geometric (directional) information in higher dimensions than wavelets can. The basic idea is to replace a scaling factor $a>0$ with an \emph{anisotropic scaling matrix}

\[
A_a^\alpha = \left( \begin{array}{cc} a & 0 \\ 0 & a^\alpha \end{array} \right),
\]

which treats the two dimensions differently and whose degree of anisotropy can be controlled via the parameter $\alpha\in[0,1]$ (for more information on so-called $\alpha$-molecules, see \citep{GKKS2014}). Naturally, anisotropic scaling requires the introduction of a third degree of freedom besides scaling and translating, namely a means to change the preferred orientation of an anisotropic two-dimensional function. The first idea which comes to mind is to add in rotations, that is to consider systems of the form 
\[
\resizebox{\hsize}{!}{$\left\{ \psi_{a,\theta,y}(x) := \psi( R_\theta A_{a}^\alpha (x - y)): a > 0, y \in \bR^2, \theta \in [0,2\pi)\right\},$}
\]
where the matrix $R_\theta$ rotates the input by $\theta$ degrees. However, rotations are not easy to implement digitally, so instead, we apply shearing. Namely, 
\begin{equation}\label{eqn:2dshear}
\resizebox{0.9\hsize}{!}{$\left\{ \psi_{a,s,y}(x) := \psi(S_sA_{a}^\alpha (x - y)): y \in \bR^2, a >0, s \in \bR\right\},$} 
\end{equation}
where $\alpha \in [0,1]$ and the \emph{shearing matrix} is defined as
\[
S_s = \left( \begin{array}{cc} 1 & s \\ 0 & 1 \end{array} \right).
\]
When we defined a wavelet system in Equation~\ref{eqn:1dwave}, $\psi_{a,y}$ meant that we had taken the generating function $\psi: \bR \rightarrow \bC$, dilated it by $a > 0$ and then translated it by $y \in \bR$.  For the \emph{shearlet system} defined in Equation~\ref{eqn:2dshear}, we start with a function $\psi: \bR^2 \rightarrow \bC$ and fix the parameter of anisotropy $\alpha$.  Then $\psi_{a,s,y}$ is $\psi$ which has been sheared by parameter $s \in \bR$ (similar to rotation), anisotropically dilated by the matrix $A_a^\alpha$, and then translated by $y \in \bR^2$.  Thus in both systems, $y$ indicates location and $a$ scale. In the shearlet system, one also gets directional information from $s$.

\begin{figure}[h]
\centering
\begin{tabular}{ccc}
\includegraphics[width=0.26\linewidth]{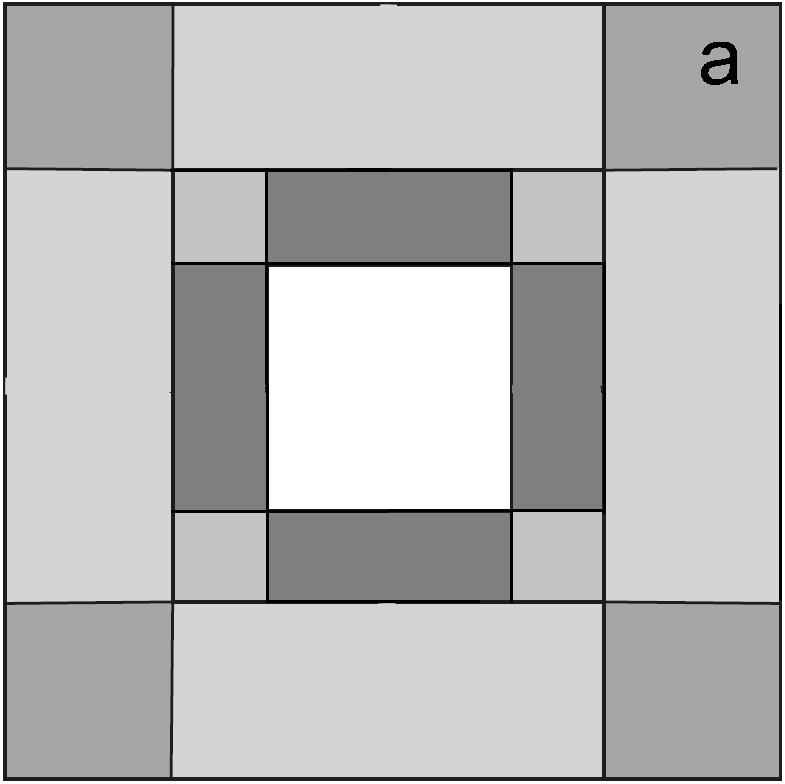} &\includegraphics[width=0.26\linewidth]{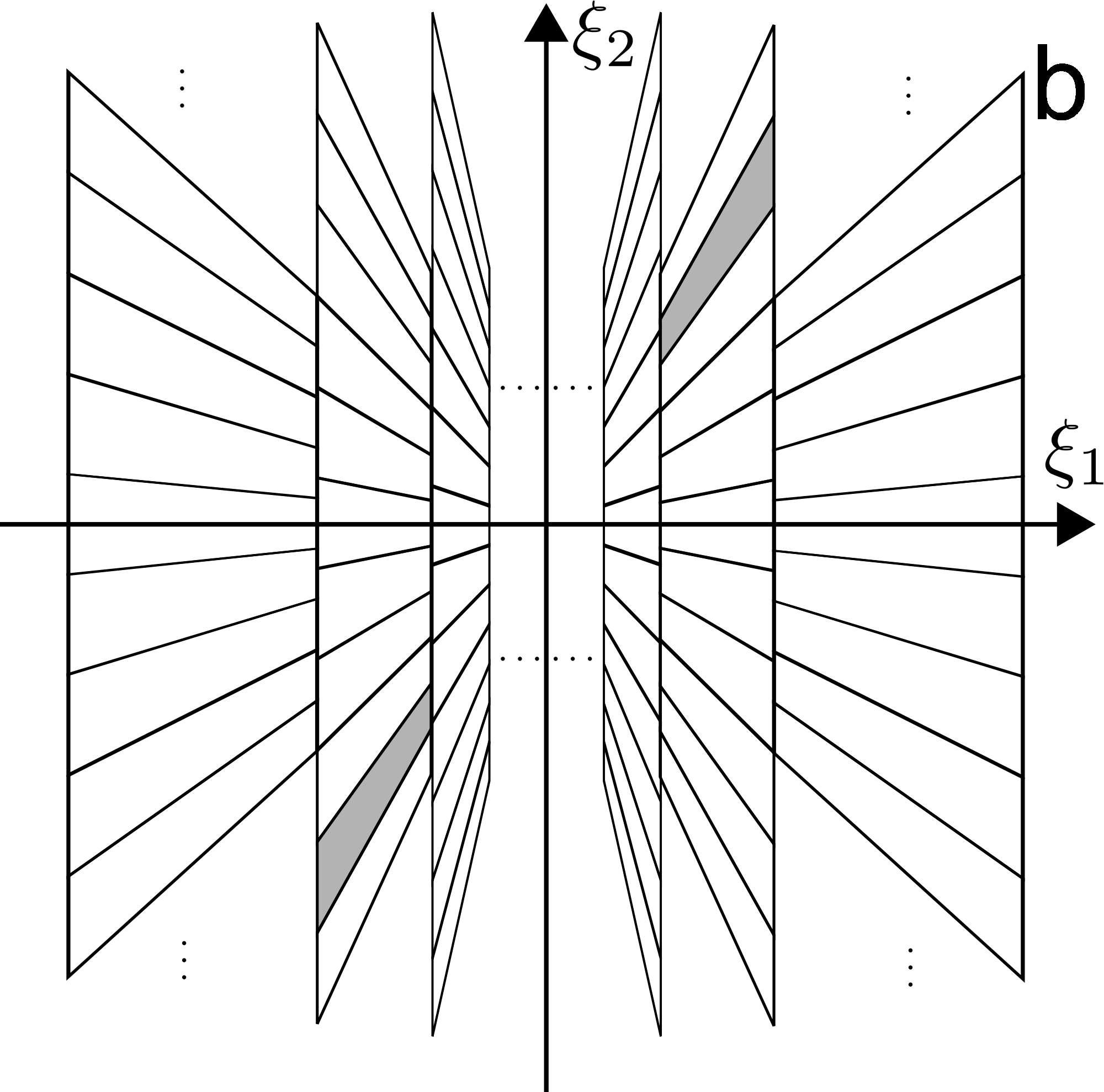} & \includegraphics[width=0.26\linewidth]{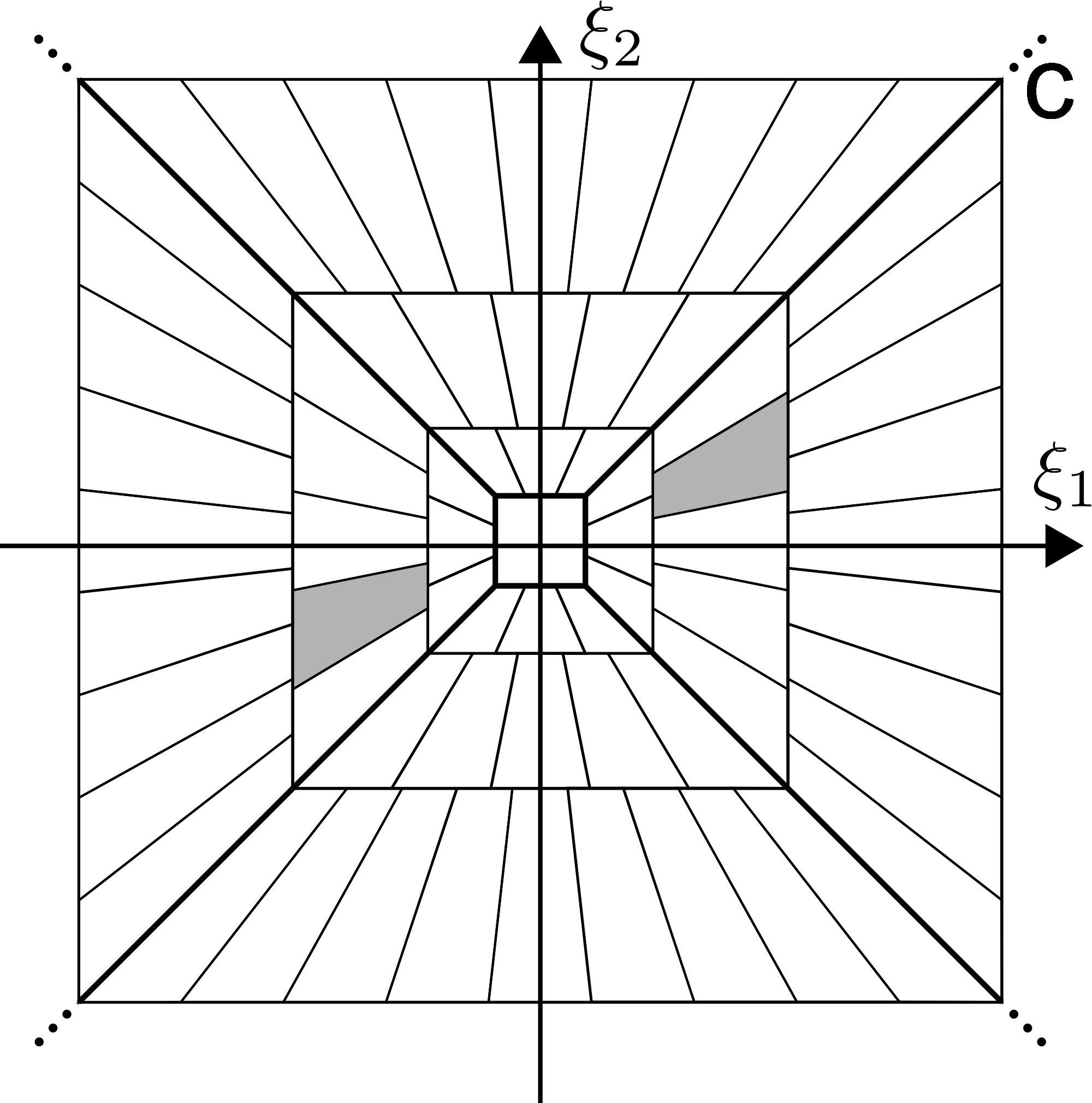}
\end{tabular}
\caption{Tiling of the frequency domain generated by the essential frequency support of (a) a wavelet system in the form of Equation~\ref{eqn:2dwave} [Meyer MRA], (b) a shearlet system in the form of Equation~\ref{eqn:2dshear}, and (c) a shearlet system with the form we use [cone-adapted]. Source: Gitta Kutyniok \& Martin Genzel, TU Berlin}
\label{fig:tiling}
\end{figure}

 In the left-hand (a) and middle (b) images of Figure~\ref{fig:tiling}, the difference between systems stemming from Equation~\ref{eqn:2dwave} and systems generated from Equation~\ref{eqn:2dshear} is depicted via the respective essential frequency supports of their elements. Note that the latter yields a polar-like decomposition. The system used in what follows is formed by gluing together two systems of the form in Equation~\ref{eqn:2dshear} generated by an even-symmetric and real-valued $\psi^{\text{(e)}}$, one oriented horizontally and one oriented vertically (see the right-hand (c) image of Figure~\ref{fig:tiling}) and then taking the even- and odd-symmetric pair formed from the Hilbert transform.  Then at each point, essentially a one-dimensional edge (resp., ridge) measure is calculated in the direction with the largest odd-symmetric (resp., even-symmetric) coefficient. More details may be found in \citep{KRKLLH}.

There is another shearlet-based edge detector which was introduced in \citep{YLEK2009}.  Their approach may be seen as a generalization of Canny which uses shearlets.  That is, they in some sense approximate the gradient of various smoothed versions of the image using a shearlet system.  The shearlet system they use is generated from a single odd function $\psi$ rather than a pair of even and odd functions $\psi^{(o)}$ and $\psi^{(e)}$ as in CoShREM.  They search for which values of $y$ give local maxima of $\absip{f}{\psi_{a,s,y}}$ for certain values of $a$ and set those to be the initial guess of the edge locations. Then they use a sophisticated technique to decrease the number of false negatives and false positives.  Since their method is built on a shearlet transform, they are also able to approximate the tangent direction of a detected edge.  Our method leverages both even- and odd-symmetric shearlet coefficients, which distinguishes our approach from theirs and also allows us to detect ridges.  Also, our implementation of the shearlet transform allows one to tune the anisotropy $\alpha$ of the scaling matrix as well as employ a finer-grained (non-dyadic) range of dilations $a$, which is not part of \citep{YLEK2009}.

\section{Results}\label{sec:results}

\subsection{Edge Detection on Mock Data}\label{sec:mock_edges}

\begin{figure*}[h]
	\centering
	
	\begin{tabular}{ccc}
		\includegraphics[width=0.3\linewidth]{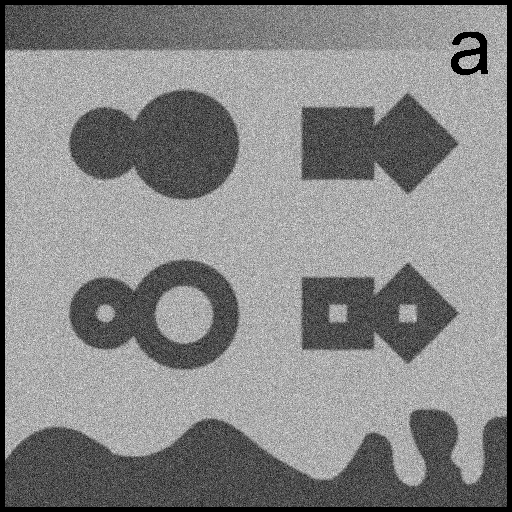}&
		\includegraphics[width=0.3\linewidth]{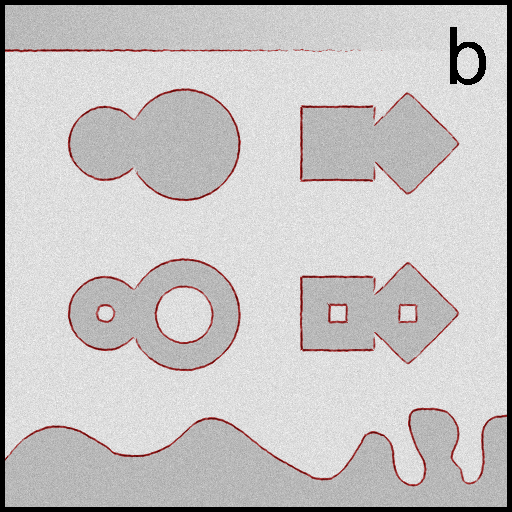}&
		\includegraphics[width=0.3\linewidth]{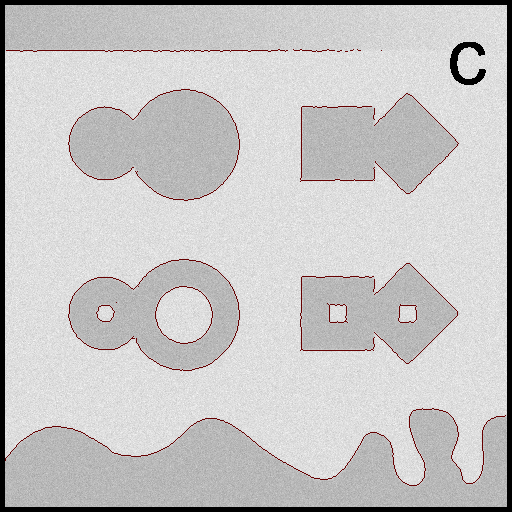}
	\end{tabular}
	
	\begin{tabular}{ccc}
		\includegraphics[width=0.3\linewidth]{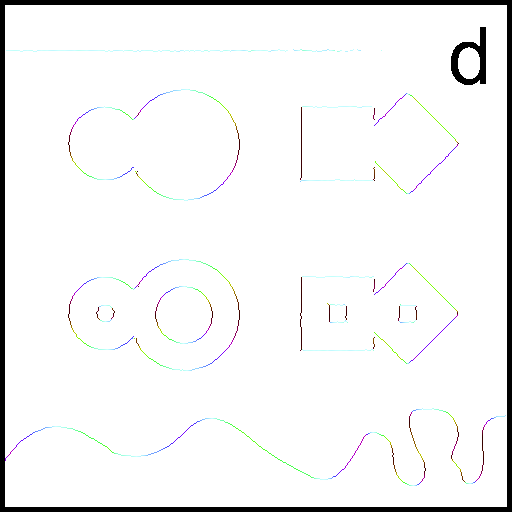}&
		\includegraphics[width=0.3\linewidth]{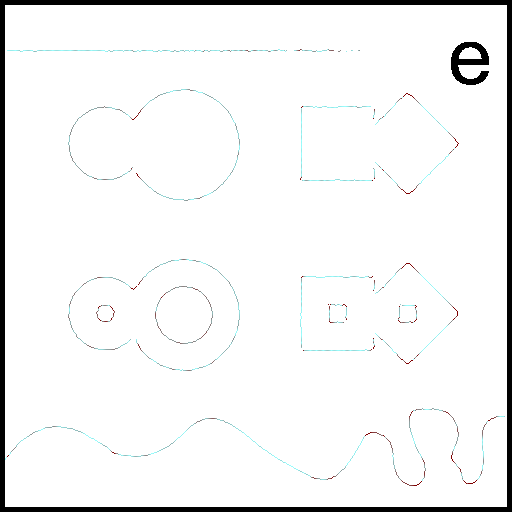}&
		\includegraphics[width=0.3\linewidth]{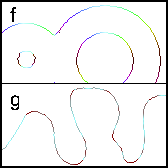}
	\end{tabular}
	\caption{Detection and analysis of edges in a noisy mock image with CoShREM. (a) Mock image perturbed with Gaussian blur ($\sigma_{\text{blur}} = 1.0$) and additive Gaussian white noise ($\sigma_{\text{noise}} = 50$). (b) Pixels colored in dark red correspond to a high value of the complex shearlet-based edge measure. For illustrative purposes, a brightened version of the processed image is shown in the background. (c) The dark red lines are obtained from thresholding and thinning the output of the shearlet-based edge measure, depicted in the previous image. (d) Color-coded estimates of the local tangent orientation, where light blue indicates a perfectly horizontal and dark red represents a perfectly vertical orientation. (e) Color-coded estimates of the local curvature, where light blue denotes zero curvature and dark red indicates a curvature greater or equal than $5^{\circ}$. (f) Enlarged section of local tangent orientation estimates. (g) Enlarged section of the local curvature estimates. }
	\label{fig:mock_edges}
\end{figure*}

We first demonstrate the applicability of CoShREM by processing a grayscale mock image containing structures typically occurring in experimentally obtained flame data which has been corrupted in various ways.  Figure~\ref{fig:mock_edges} contains the visual results of CoShREM being applied in one such experiment. Figure~\ref{fig:mock_edges}(a) is the mock image after corruption by Gaussian blur ($\sigma_{\text{blur}} = 1.0$) and additive Gaussian white noise ($\sigma_{\text{noise}} = 50$). Alongside the original (Figure~\ref{fig:mock_edges}(b)) and thresholded, thinned (Figure~\ref{fig:mock_edges}(c)) values of the complex shearlet-based edge measure, estimates of the local tangent orientation and the local curvature are plotted in Figure~\ref{fig:mock_edges}(d)--(g).

\renewcommand{\arraystretch}{3}
\renewcommand{\arraystretch}{1}

\

In the experiments, we compare the newly proposed measure to a number of established methods such as the Canny edge detector \citep{Can1986}, the Sobel edge detector \citep{SF68}, and the phase congruency measure \citep{Kov1999,Kov2000} as well as another shearlet-based edge detector developed by Yi et al. in 2009 \citep{YLEK2009}. This comparison is carried out visually but also numerically by computing Pratt's figure of merit (PFOM) \citep{AbPr79} on the obtained results\footnote{While the corresponding ground-truth is often created applying one of the edge detectors on the noiseless image, in this case, it was handmade by the authors to prevent favoring  a particular method.}. To put an emphasis on testing the stability of CoShREM, two different types of experiments were performed on the mock image.  First, in line with traditional image processing literature, the image analyzed in Figure~\ref{fig:mock_edges} was perturbed by four levels of Gaussian blur  (i.e. convolution with a Gaussian filter kernel) and five levels of additive Gaussian white noise for a total of 20 different levels of corruption before CoShREM and other edge detection algorithms were applied. Second, there are a number of different sources of noise (electronic noise, photon noise/shot noise; readout noise, etc) in combustion diagnostics, and their accumulated effect on the image depends on the camera model, the laser source, the flame condition, the imaging optics etc. \citep{PBLL06,FKL05,SHDH94,MPH93}.  Therefore, the assumption of an overall Gaussian noise distribution is reasonable.   However, for completeness we have run a comparison of the edge detection algorithms on the mock data which has been corrupted by five different levels of additive Gaussian noise and then Poisson/shot noise. For each of these test images, the parameters configuring the various edge detection algorithms remained fixed. However, for each method, the set of parameters was chosen such that both visually and with respect to the PFOM metric, the maximal error was minimized across all levels of noise. To highlight the importance of carefully choosing fitting parameters for each edge detector, the Canny edge detector applied with its default parameters in MATLAB was also included in the comparison.

\

In the case of the Canny and the Sobel edge detectors, the implementations provided by the MATLAB Image Processing Toolbox (version 9.2) were applied. The software used to compute the phase congruency measure can be downloaded from Peter Kovesi's homepage \citep{KovONLINE} while an implementation of the other shearlet-based edge detector was kindly provided by the authors of \citep{YLEK2009}. For the latter, pre-processing with a Gaussian smoothing filter was added such that the algorithm could also handle more severe levels of noise.

\

\begin{figure*}[h]
	\centering
	\begin{tabular}{ccc}
	\includegraphics[width=0.3\linewidth]{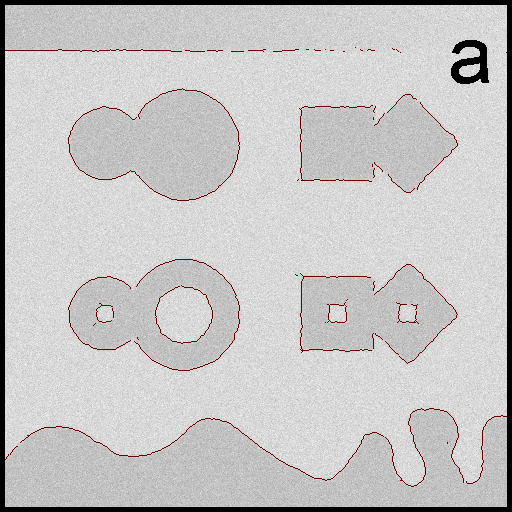}&
    \includegraphics[width=0.3\linewidth]{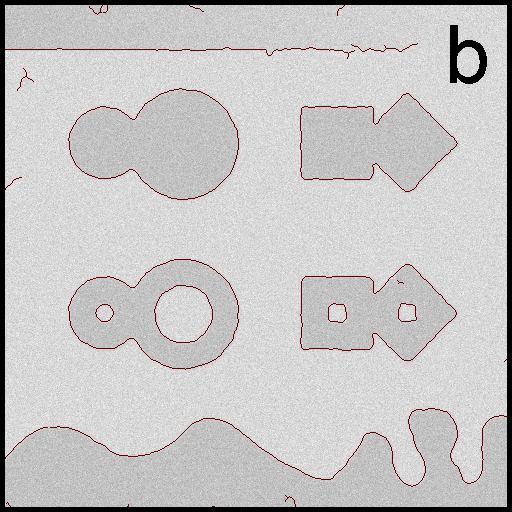}&
	\includegraphics[width=0.3\linewidth]{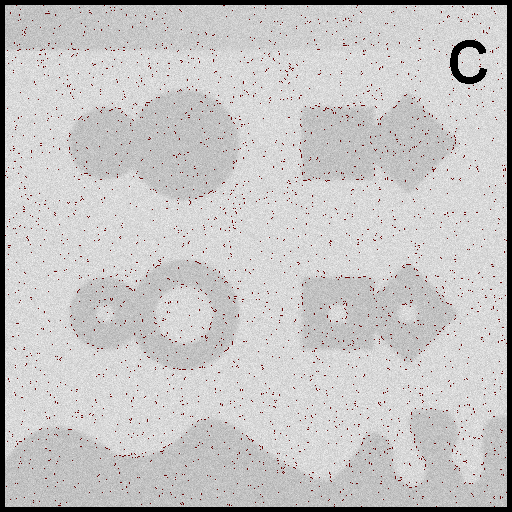}\\	
	\includegraphics[width=0.3\linewidth]{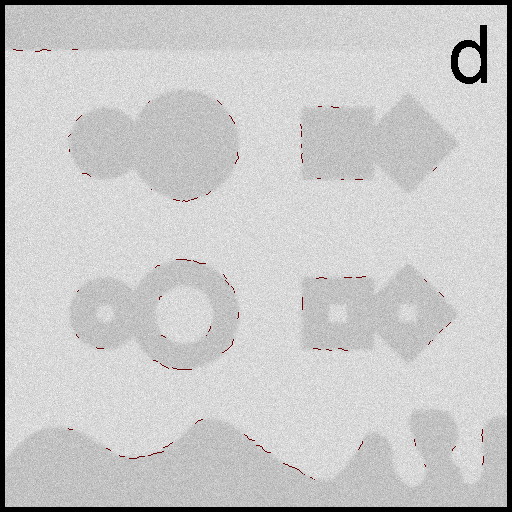}&
	\includegraphics[width=0.3\linewidth]{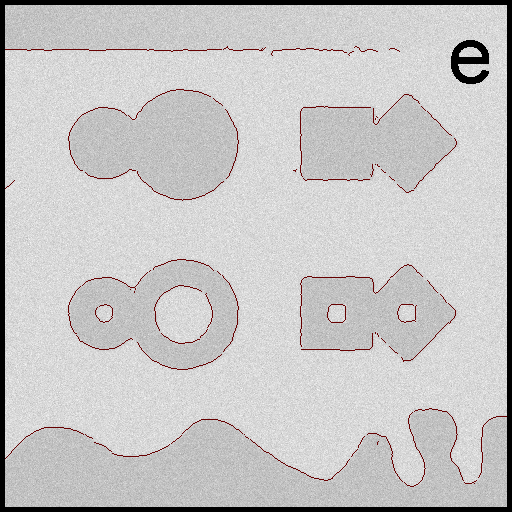}&
	\includegraphics[width=0.3\linewidth]{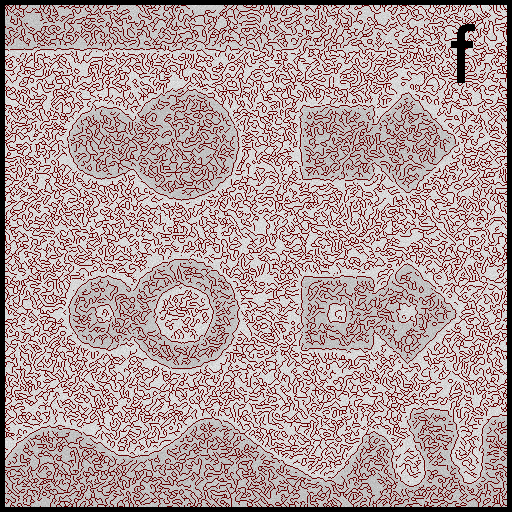}
	\end{tabular}
	\caption{Visual comparison of different edge detection algorithms. The processed image was perturbed with Gaussian blur ($\sigma_{\text{blur}} = 1.5$) and additive Gaussian white noise ($\sigma_{\text{noise}} = 100$). The displayed results were obtained from (a) CoShREM, (b) the Canny edge detector, (c) the Sobel edge detector, (d) the phase congruency measure, (e) Yi et al.'s shearlet edge detector, and (f) the Canny edge detector with its default configuration in MATLAB. The PFOM values corresponding to the results shown here can be found in the last column of Table~\ref{table:mock_edges_comparison}. For illustrative purposes, brightened versions of the processed images are shown in the background.}
	\label{fig:mock_edges_comparison_noisy}
\end{figure*}

All results of the success of the edge detection algorithms applied to the images corrupted by Gaussian blur and additive Gaussian noise with respect to the PFOM metric are compiled in Table~\ref{table:mock_edges_comparison} while detection results for all algorithms for the noisiest case are plotted in Figure~\ref{fig:mock_edges_comparison_noisy}.  The PFOM values of the different methods applied to the mock image perturbed by Gaussian noise followed by Poisson noise are in Table~\ref{table:mock_edges_comparison_poisson}.  Under all noise and blurring regimes, CoShREM yields the best PFOM values.

\

\begin{figure*}[h]
	\centering
	\begin{tabular}{ccc}
	\includegraphics[width=0.3\linewidth]{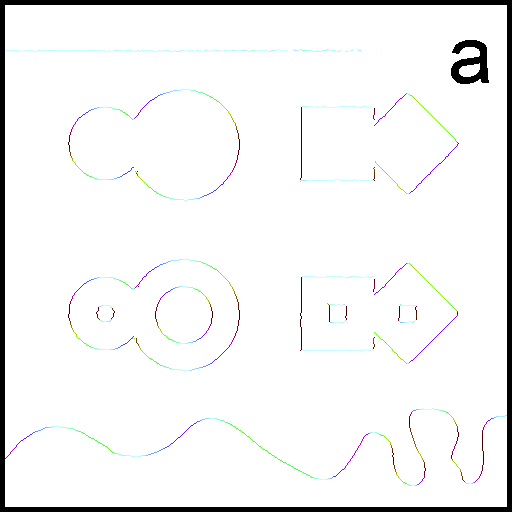}&
    \includegraphics[width=0.3\linewidth]{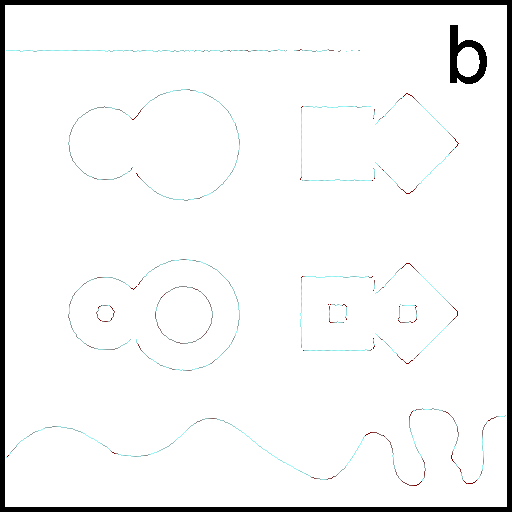}&
	\includegraphics[width=0.3\linewidth]{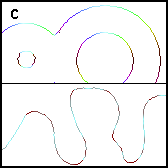}\\	
	\includegraphics[width=0.3\linewidth]{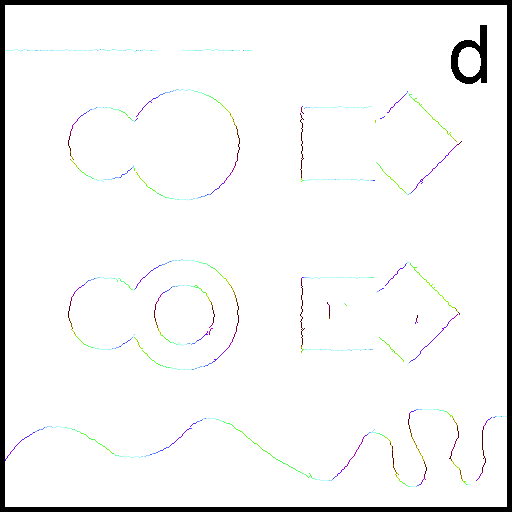}&
	\includegraphics[width=0.3\linewidth]{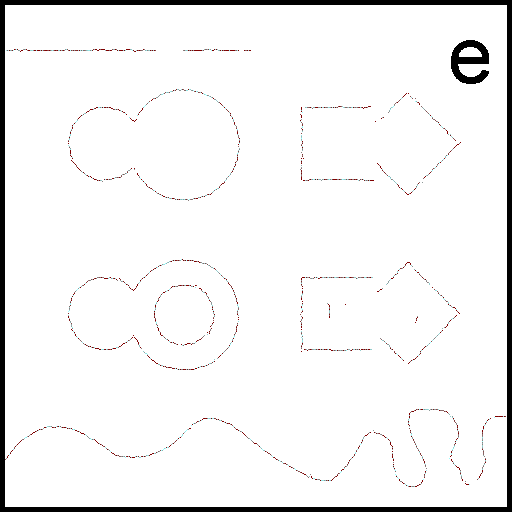}&
	\includegraphics[width=0.3\linewidth]{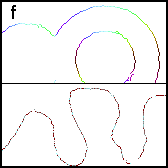}\\	
	\includegraphics[width=0.3\linewidth]{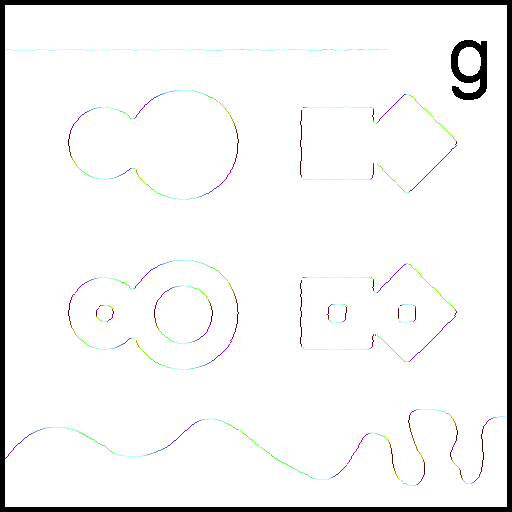}&
	\includegraphics[width=0.3\linewidth]{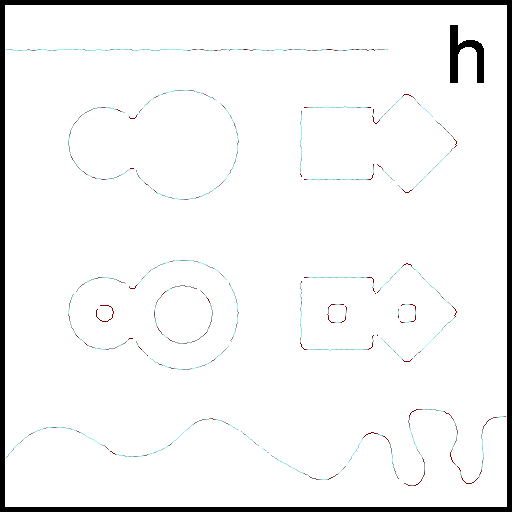}&
	\includegraphics[width=0.3\linewidth]{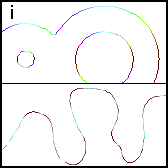}
	\end{tabular}
	\caption{Visual comparison of tangent orientation and curvature estimates obtained from different algorithms. The processed image is the same as in figure~\ref{fig:mock_edges} ($\sigma_{\text{blur}} = 1.0$, $\sigma_{\text{noise}} = 50$). Light blue indicates a perfectly horizontal and dark red represents a perfectly vertical orientation in the first column while light blue denotes zero curvature and dark red indicates a curvature greater or equal than $5^{\circ}$ in the middle column. The final column shows enlarged sections of the preceding images. (a,b,c) Results obtained from CoShREM. (d,e,f) Results obtained from the phase congruency measure. (g,h,i) Results obtained from Yi et al.'s shearlet edge detector.}
	\label{fig:mock_edges_orientation_curvature_comparison}
\end{figure*}

While the Canny and Sobel edge detectors are based on approximations of the local gradients in horizontal and vertical direction, the three other methods considered apply analyzing elements associated with numerous orientations, thereby automatically yielding more or less refined estimates of the local tangent orientation. A visual comparison of the approximated local orientation and curvature of the edges in the distorted mock image from Figure~\ref{fig:mock_edges} is depicted in Figure~\ref{fig:mock_edges_orientation_curvature_comparison}. Please note that the computation of the local tangent orientations differs in all three applied edge detectors. However, all curvature estimates were made by computing the central difference with respect to the local tangent orientation.

\setlength{\tabcolsep}{2.5pt}
\begin{table}[h]
\begin{scriptsize}
\centering
\begin{tabular}{|r|ccccc|}
\hline
&\multicolumn{5}{c|}{$\sigma_{\text{blur}} = 0.0$}\\
\cline{2-6}
 & $\sigma_{\text{noise}}=0$ &$20$&$50$&$80$&$100$ \\
\hline
CoShREM &0.97&0.96&0.93&0.92&0.91\\

Canny & 0.90 & 0.90 & 0.88 & 0.88 & 0.88\\

Sobel & 0.93 & 0.92 & 0.89 & 0.70 & 0.44\\

Phase congruency & 0.95 & 0.94 & 0.90 & 0.78 & 0.72\\

Yi et al. shearlet edge detector & 0.88 & 0.88 & 0.88 & 0.87 & 0.88\\

Canny (default parameters) & 0.92 & 0.11 & 0.10 & 0.11 & 0.11\\
\hline
\end{tabular}
\begin{tabular}{|r|ccccc|}
\hline
&\multicolumn{5}{c|}{$\sigma_{\text{blur}} = 0.5$} \\
\cline{2-6}
 & $\sigma_{\text{noise}}=0$ & $20$ & $50$ & $80$ & $100$ \\
\hline
CoShREM  & 0.97 & 0.96 & 0.93 & 0.92 & 0.91 \\

Canny & 0.90 & 0.90 & 0.88 & 0.89 & 0.88 \\

Sobel & 0.93 & 0.91 & 0.89 & 0.65 & 0.43 \\

Phase congruency &  0.95 & 0.94 & 0.88 & 0.75 & 0.62 \\

Yi et al. shearlet edge detector & 0.88 & 0.89 & 0.88 & 0.88 & 0.87 \\

Canny (default parameters) & 0.92 & 0.11 & 0.10 & 0.11 & 0.11\\
\hline
\end{tabular}
\begin{tabular}{|r|ccccc|}
\hline
&\multicolumn{5}{c|}{$\sigma_{\text{blur}} = 1.0$} \\
\cline{2-6}
 & $\sigma_{\text{noise}}=0$ & $20$ & $50$ & $80$ & $100$ \\
\hline
CoShREM & 0.96 & 0.95 & 0.94 & 0.92 & 0.91  \\

Canny & 0.89 & 0.89 & 0.89 & 0.89 & 0.89  \\

Sobel & 0.91 & 0.90 & 0.76 & 0.44 & 0.33 \\

Phase congruency & 0.94 & 0.92 & 0.83 & 0.60 & 0.39 \\

Yi et al. shearlet edge detector & 0.88 & 0.88 & 0.88 & 0.87 & 0.85 \\

Canny (default parameters) & 0.92 & 0.10 & 0.10 & 0.10 & 0.11\\
\hline
\end{tabular}
\\

\begin{tabular}{|r|ccccc|}
\hline
&\multicolumn{5}{c|}{$\sigma_{\text{blur}} = 1.5$}\\
\cline{2-6}
 & $\sigma_{\text{noise}}=0$ & $20$ & $50$ & $80$ & $100$ \\
\hline
CoShREM &  0.95 & 0.94 & 0.93 & 0.90 & 0.89 \\

Canny &0.89 & 0.89 & 0.88 & 0.87 & 0.87 \\

Sobel & 0.89 & 0.93 & 0.47 & 0.32 & 0.27 \\

Phase congruency & 0.89 & 0.80 & 0.72 & 0.33 & 0.16 \\

Yi et al. shearlet edge detector & 0.87 & 0.88 & 0.87 & 0.86 & 0.86 \\

Canny (default parameters) &0.91 & 0.10 & 0.10 & 0.10 & 0.11 \\
\hline
\end{tabular}
\caption{Numerical comparison of CoShREM with five other edge detectors. The table shows PFOM values for all considered algorithms and a total of 20 differently distorted versions of the mock image shown in Figure~\ref{fig:mock_edges}, where 1.0 would indicate a perfect reproduction of the ground-truth. For each algorithm, parameters remained fixed for all test images but were carefully optimized such that the maximal error was minimized across all levels of noise. The binary ground-truth was drawn from hand and consisted of minimally connected lines (i.e. with the exception of intersections, each pixel with value $1$ has at most two neighbors with value $1$). To ensure a fair comparison, a thinning operation was applied to the binary outcome of each method. For a visual comparison of the results in the noisiest case ($\sigma_{\text{blur}} = 1.5$, $\sigma_{\text{noise}} = 100$), see Figure~\ref{fig:mock_edges_comparison_noisy}.}
\label{table:mock_edges_comparison}
\end{scriptsize}
\end{table}

\begin{table}[h]
	\begin{scriptsize}
		\centering
		\begin{tabular}{|r|ccccc|}
			\hline
			&\multicolumn{5}{c|}{Poisson and Gaussian noise}\\
			\cline{2-6}
			& $\sigma_{\text{noise}}=0$ &$20$&$50$&$80$&$100$ \\
			\hline
			CoShREM & 0.95 & 0.94 & 0.91 & 0.89 & 0.83\\			
			Canny & 0.90 & 0.89 & 0.87 & 0.50 & 0.28\\			
			Sobel & 0.65 & 0.49 & 0.36 & 0.22 & 0.24\\			
			Phase congruency & 0.93 & 0.90 & 0.70 & 0.24 & 0.00\\			
			Yi et al. shearlet edge detector & 0.89 & 0.88 & 0.86 & 0.78 & 0.65\\			
			Canny (default parameters) & 0.18 & 0.11 & 0.11 & 0.11 & 0.12\\
			\hline
		\end{tabular}
		\caption{Numerical comparison of the stability under additional Poisson noise. The table shows PFOM values for all six considered edge detectors and a total of five differently distorted versions of the mock image shown in Figure~\ref{fig:mock_edges}(a), where 1.0 would indicate a perfect reproduction of the ground-truth. The test images were first perturbed with five different levels of additive Gaussian noise. Then, each pixel was resampled from a Poisson distribution with an expectancy of one tenth of the original pixel value. Finally, the values of the thereby obtained grayscale image were rescaled by a factor of $10$. To test the stability with respect to this kind of shot noise, the same parameters as in Table~\ref{table:mock_edges_comparison} were used for all algorithms. }
		\label{table:mock_edges_comparison_poisson}
	\end{scriptsize}
\end{table}

\setlength{\tabcolsep}{6pt}
  \renewcommand{\arraystretch}{3}
  
\subsection{Ridge Detection on Mock Data}\label{sec:mock_lines}

\begin{figure*}[h]
	\centering
	\begin{tabular}{ccc}
		\includegraphics[width=0.3\linewidth]{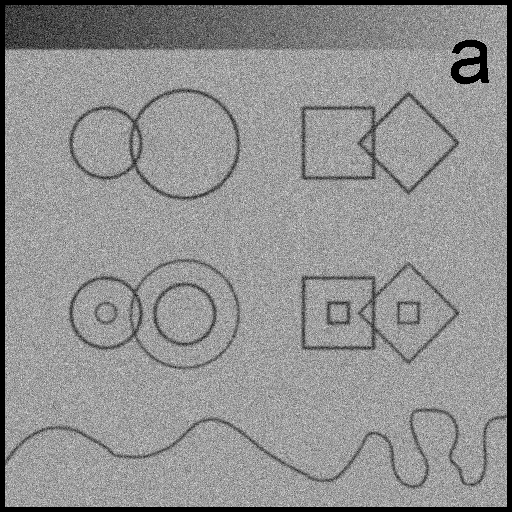}&
		\includegraphics[width=0.3\linewidth]{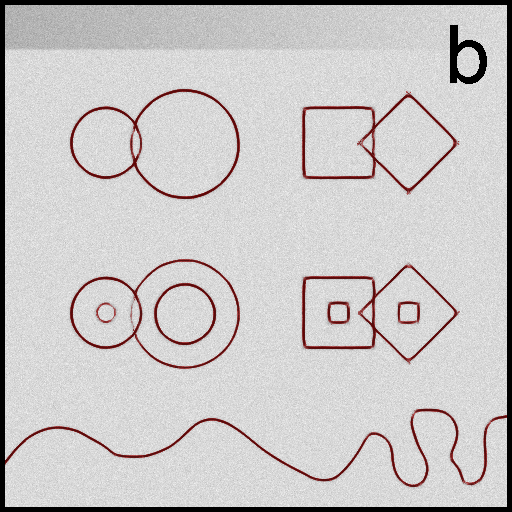}&
		\includegraphics[width=0.3\linewidth]{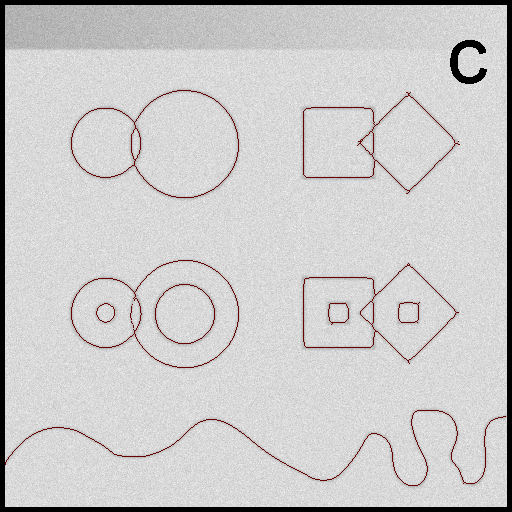}
	\end{tabular}
	\begin{tabular}{ccc}
		\includegraphics[width=0.3\linewidth]{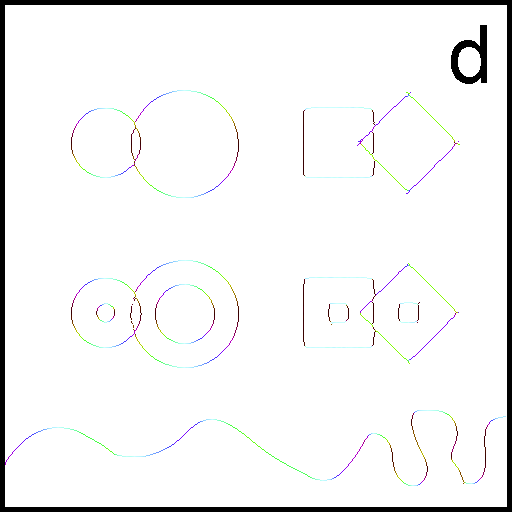}&
		\includegraphics[width=0.3\linewidth]{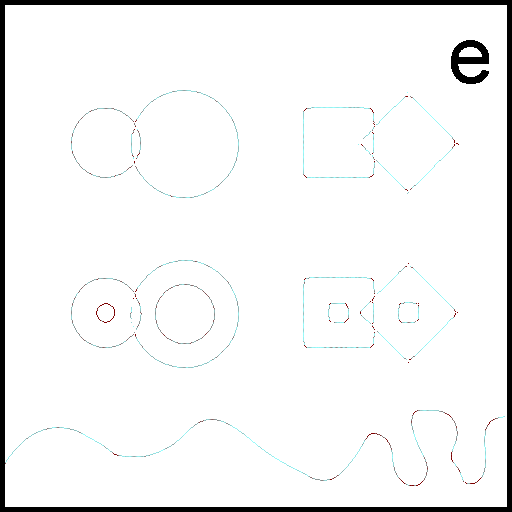}&
		\includegraphics[width=0.3\linewidth]{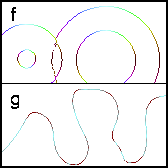}
	\end{tabular}
	\caption{Detection and analysis of ridges in a noisy mock image with CoShREM. (a) Mock image perturbed with Gaussian blur ($\sigma_{\text{blur}} = 1.0$) and additive Gaussian white noise ($\sigma_{\text{noise}} = 50$). (b) Pixels colored in dark red correspond to a high value of the complex shearlet-based ridge measure. For illustrative purposes, a brightened version of the processed image is shown in the background. (c) The red lines are obtained from thresholding and thinning the output of CoShREM, depicted in the previous image. (d) Color-coded estimates of the local tangent orientation, where light blue indicates a perfectly horizontal and dark red represents a perfectly vertical orientation. (e) Color-coded estimates of the local curvature, where light blue denotes zero curvature and dark red indicates a curvature greater or equal than $5^{\circ}$. (f) Enlarged section of local tangent orientation estimates. (g) Enlarged section of the local curvature estimates.}
	\label{fig:mock_lines}
\end{figure*}

Just as in the edge detection case, we demonstrate the applicability of the complex shearlet-based ridge measure by processing a distorted grayscale mock image (see Figure~\ref{fig:mock_lines}).

\renewcommand{\arraystretch}{1}

\

Out of the five algorithms considered in Section~\ref{sec:mock_edges} only CoShREM and the phase congruency measure are capable of detecting ridges as coherent structures (see Figure~\ref{fig:ridgevsedgemeasure}). To the other methods, a ridge just looks like a thin homogeneous region bounded by two edges. Hence, instead of a single ridge, two edges would be detected. This behavior is highly unfeasible for the detection of flame fronts as these edges will get increasingly weak and hidden by noise as ridges get thinner. Furthermore, merging two detected edges to calculate the location of a single ridge would require an additional post-processing step that again is non-trivial and could also destroy finer structure.

\

\begin{figure*}[h]
	\centering
	\begin{tabular}{ccc}
	\includegraphics[width=0.3\linewidth]{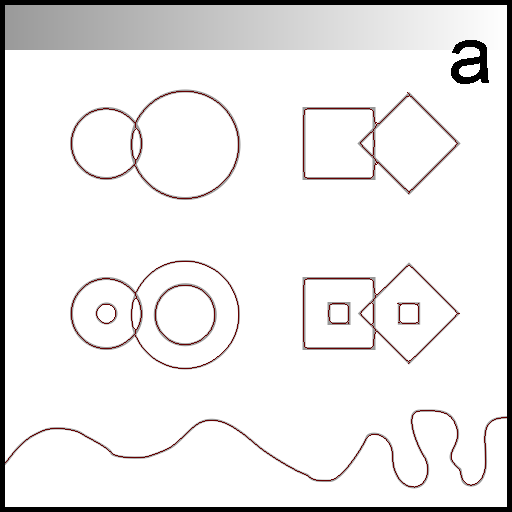}&
    \includegraphics[width=0.3\linewidth]{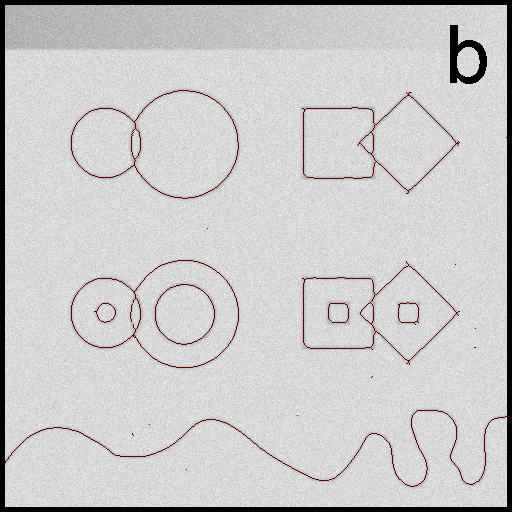}&
	\includegraphics[width=0.3\linewidth]{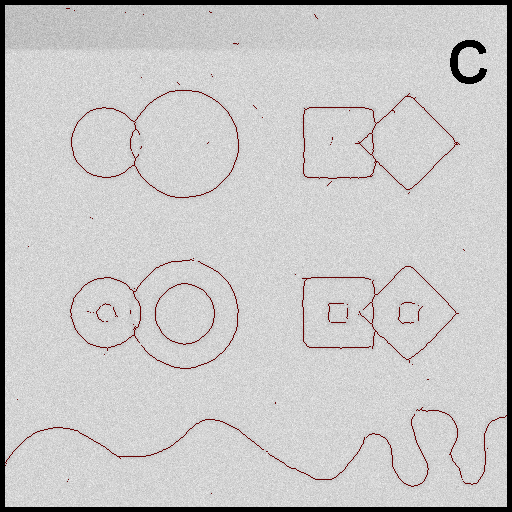}\\	
	\includegraphics[width=0.3\linewidth]{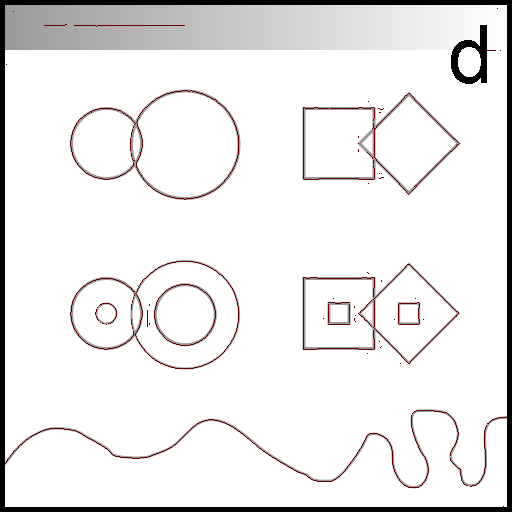}&
    \includegraphics[width=0.3\linewidth]{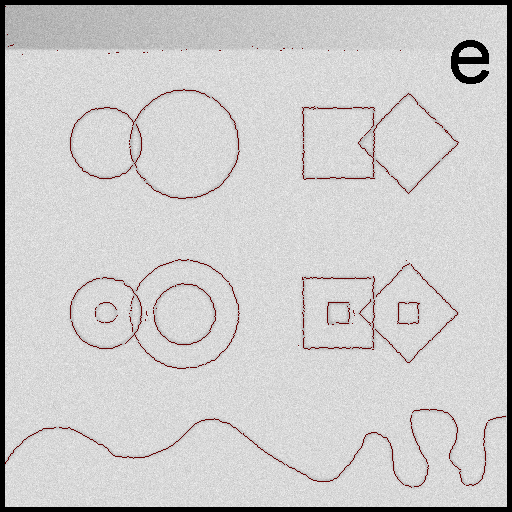}&
	\includegraphics[width=0.3\linewidth]{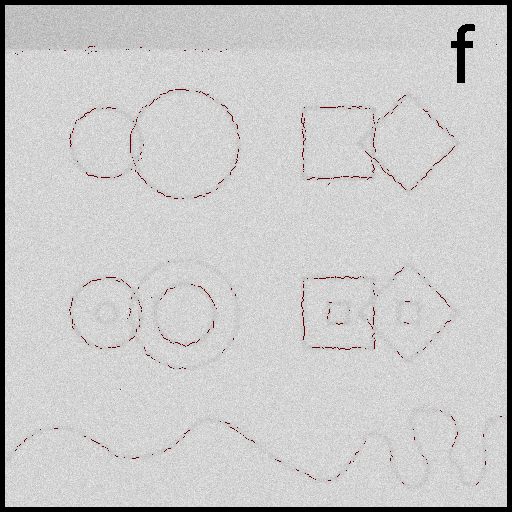}
	\end{tabular}
	\caption{Visual comparison of CoShREM (a,b,c) and the phase congruency-based ridge detector (d,e,f). The PFOM values corresponding to the results shown here can be found in the first ($\sigma_{\text{blur}} = 0$, $\sigma_{\text{noise}} = 0$), thirteenth ($\sigma_{\text{blur}} = 1.0$, $\sigma_{\text{noise}} = 50$), and last ($\sigma_{\text{blur}} = 1.5$, $\sigma_{\text{noise}} = 100$) column of Table~\ref{table:mock_lines_comparison}. For illustrative purposes, brightened versions of the processed images are shown in the background.}
	\label{fig:mock_lines_comparison}
\end{figure*}

Again, the robustness of CoShREM is subjected to a detailed analysis by processing a total of 20 differently distorted mock images with a fixed set of parameters. This time however, for the aforementioned reasons, only the phase congruency measure is included in this analysis, which again is based on a visual comparison and a numerical comparison via the PFOM metric. The numerical results are summarized in Table~\ref{table:mock_lines_comparison}. A visual comparison of the detection results of the complex shearlet-based measure and the phase congruency measure for three differently perturbed test images is provided in Figure~\ref{fig:mock_lines_comparison}.

\

\begin{figure*}[h]
	\centering
	\begin{tabular}{ccc}
		\includegraphics[width=0.3\linewidth]{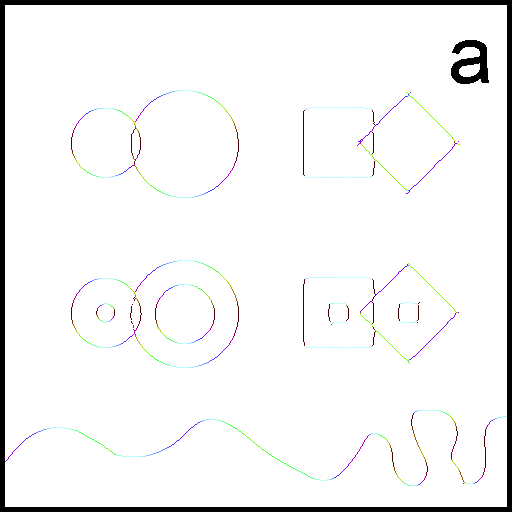}&	
		\includegraphics[width=0.3\linewidth]{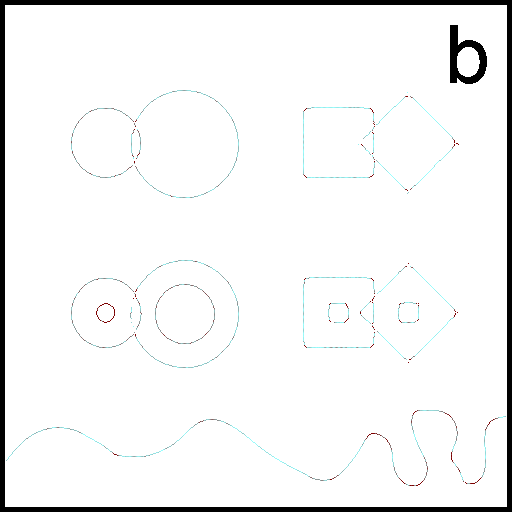}&
		\includegraphics[width=0.3\linewidth]{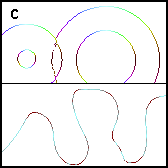}\\
	    \includegraphics[width=0.3\linewidth]{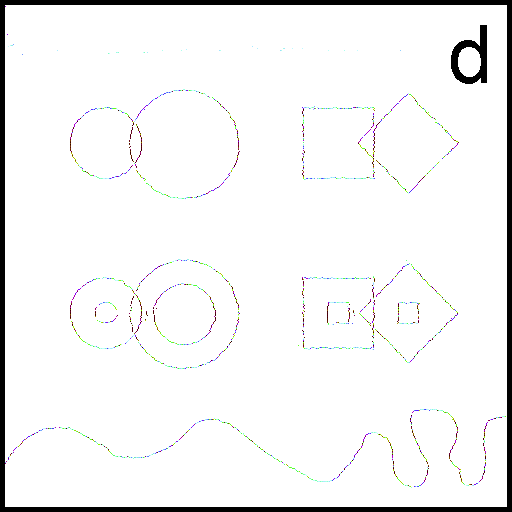}&
	    \includegraphics[width=0.3\linewidth]{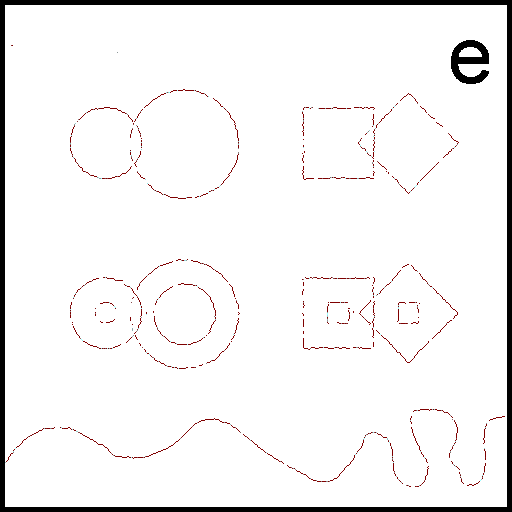}&
	    \includegraphics[width=0.3\linewidth]{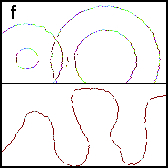}
	\end{tabular}
	\caption{Visual comparison of tangent orientation and curvature estimates obtained from CoShREM (a,b,c) and the phase congruency measure (d,e,f). The processed image is the same as in figure Figure~\ref{fig:mock_lines} ($\sigma_{\text{blur}} = 1.0$, $\sigma_{\text{noise}} = 50$). Light blue indicates a perfectly horizontal and dark red represents a perfectly vertical orientation in the first column while light blue denotes zero curvature and dark red indicates a curvature greater or equal than $5^{\circ}$ in the middle column. The final column shows enlarged sections of the preceding images.}
	\label{fig:mock_lines_orientation_curvature_comparison}
\end{figure*}

Finally, a visual comparison of the tangent orientation and curvature estimates obtained from the complex shearlet-based ridge measure and the phase congruency measure for the distorted mock image from Figure~\ref{fig:mock_lines} can be found in Figure~\ref{fig:mock_lines_orientation_curvature_comparison}.

\setlength{\tabcolsep}{2.5pt}
\begin{table}[h]
\begin{scriptsize}
\centering
\begin{tabular}{|r|ccccc|}
\hline
&\multicolumn{5}{c|}{$\sigma_{\text{blur}} = 0.0$} \\
\cline{2-6}
 & $\sigma_{\text{noise}}=0$ & $20$ & $50$ & $80$ & $100$\\
\hline
CoShREM & 0.94 & 0.95 & 0.95 & 0.93 & 0.93 \\
Phase congruency & 0.90 & 0.94 & 0.94 & 0.93 & 0.93 \\
\hline
\end{tabular}
\begin{tabular}{|r|ccccc|}
\hline
 & \multicolumn{5}{c|}{$\sigma_{\text{blur}} = 0.5$}\\
\cline{2-6}
 & $\sigma_{\text{noise}}=0$ & $20$ & $50$ & $80$ & $100$\\
\hline
CoShREM & 0.93 & 0.93 & 0.92 & 0.89 & 0.89 \\
Phase congruency & 0.88 & 0.92 & 0.93 & 0.92 & 0.88 \\
\hline
\end{tabular}
\begin{tabular}{|r|ccccc|}
\hline
&\multicolumn{5}{c|}{$\sigma_{\text{blur}} = 1.0$} \\
\cline{2-6}
 & $\sigma_{\text{noise}}=0$ & $20$ & $50$ & $80$ & $100$  \\
\hline
CoShREM & 0.94 & 0.92 & 0.92 & 0.90 & 0.86 \\
Phase congruency & 0.88 & 0.92 & 0.92 & 0.88 & 0.81 \\
\hline
\end{tabular}
\begin{tabular}{|r|ccccc|}
\hline
& \multicolumn{5}{c|}{$\sigma_{\text{blur}} = 1.5$}\\
\cline{2-6}
 & $\sigma_{\text{noise}}=0$ & $20$ & $50$ & $80$ & $100$ \\
\hline
CoShREM & 0.93 & 0.92 & 0.91 & 0.88 & 0.88 \\
Phase congruency & 0.88 & 0.91 & 0.89 & 0.74 & 0.37 \\
\hline
\end{tabular}
\caption{Numerical comparison of CoShREM and the phase congruency-based ridge detector. The table shows again PFOM values, where 1.0 would indicate a perfect reproduction of the ground-truth. Both methods were applied to a total of 20 differently distorted versions of the mock image shown in Figure~\ref{fig:mock_lines}. For both algorithms, parameters remained fixed for all test images but were carefully optimized such that the maximal error was minimized across all levels of noise. The binary ground-truth was drawn from hand and consisted of minimally connected lines (i.e. with the exception of intersections, each pixel with value $1$ has at most two neighbors with value $1$). To ensure a fair comparison, a thinning operation was applied to the binary outcome of each method. For a visual comparison for three differently distorted images, see Figure~\ref{fig:mock_lines_comparison}.}
\label{table:mock_lines_comparison}
\end{scriptsize}
\end{table}
\setlength{\tabcolsep}{6pt}

\
\renewcommand{\arraystretch}{3}

\subsection{Edge and Ridge Detection on PLIF Images}\label{sec:real}

\begin{figure*}[h]
	\centering
	\begin{tabular}{ccccc}
	\includegraphics[width=0.180\linewidth]{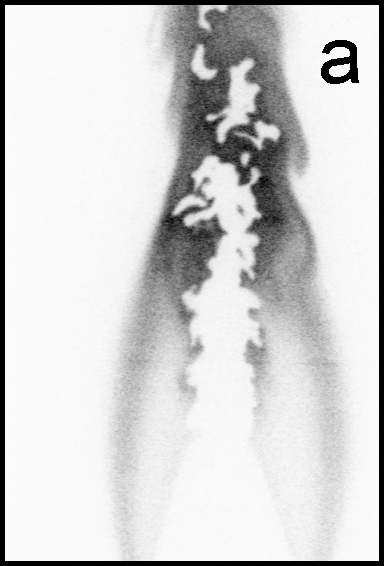}&
	\includegraphics[width=0.180\linewidth]{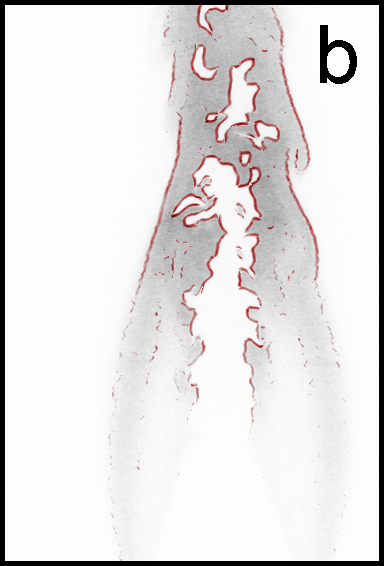}&
	\includegraphics[width=0.180\linewidth]{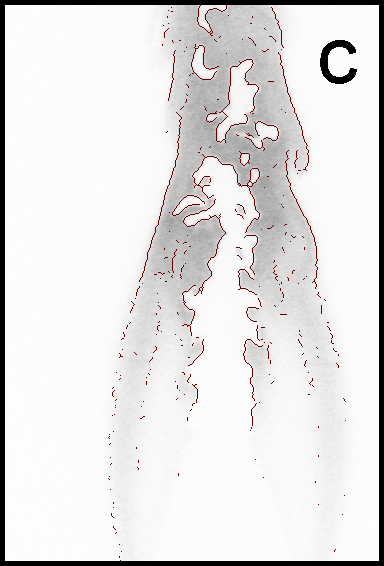}&
	\includegraphics[width=0.180\linewidth]{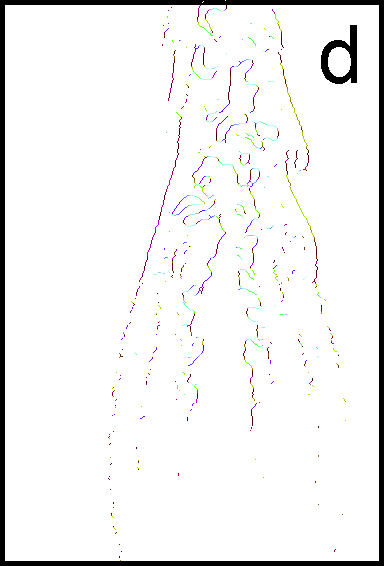}&
	\includegraphics[width=0.180\linewidth]{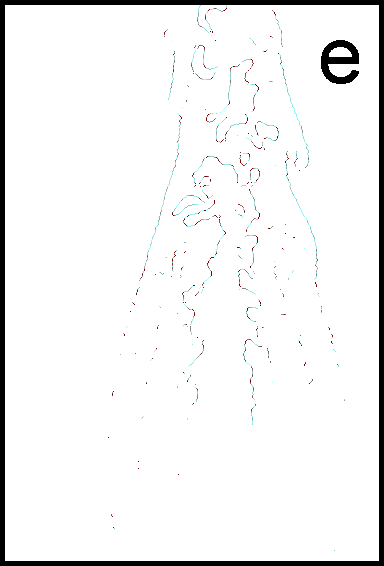}\\	\includegraphics[width=0.180\linewidth]{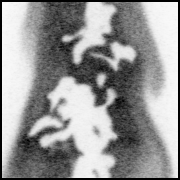}&
	\includegraphics[width=0.180\linewidth]{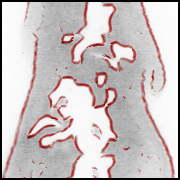}&
	\includegraphics[width=0.180\linewidth]{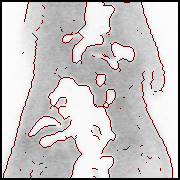}&
	\includegraphics[width=0.180\linewidth]{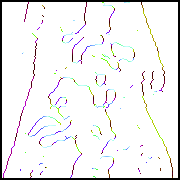}&
	\includegraphics[width=0.180\linewidth]{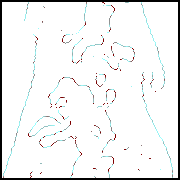}
	\end{tabular}
	\caption{Detection and analysis of flame fronts with CoShREM. (a) PLIF visualization of long-lived {OH} radicals. (b) Pixels colored in dark red correspond to a high value of CoShREM. For illustrative purposes, a brightened version of the processed image is shown in the background. (c) The red lines are obtained from thresholding and thinning the output of CoShREM, depicted in the previous image. (d) Color-coded estimates of the local tangent orientation, where light blue indicates a perfectly horizontal and dark red represents a perfectly vertical orientation. (e) Color-coded estimates of the local curvature, where light blue denotes zero curvature and dark red indicates a curvature greater or equal than $15^{\circ}$. The second row depicts enlarged sections of the images shown in the top row.}
	\label{fig:real_edges}
\end{figure*}

\begin{figure*}[h]
	\centering
	\begin{tabular}{ccccc}
		\includegraphics[width=0.180\linewidth]{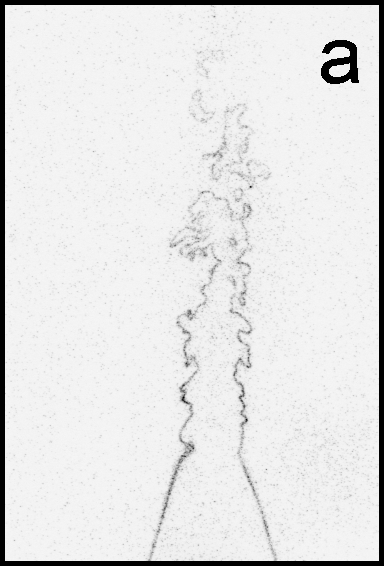}&
		\includegraphics[width=0.180\linewidth]{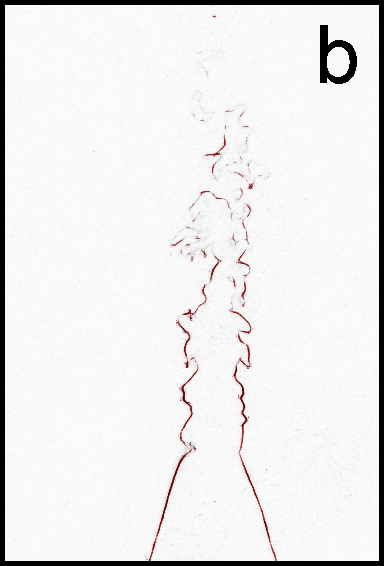}&
		\includegraphics[width=0.180\linewidth]{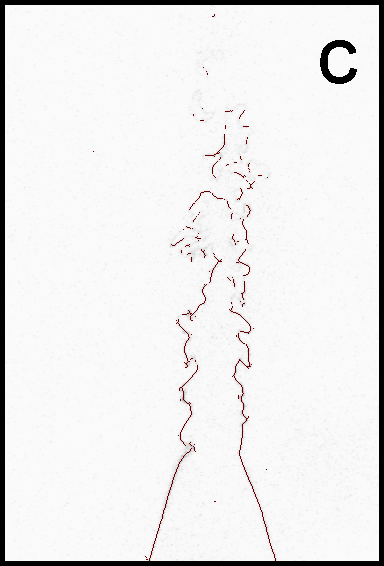}&
		\includegraphics[width=0.180\linewidth]{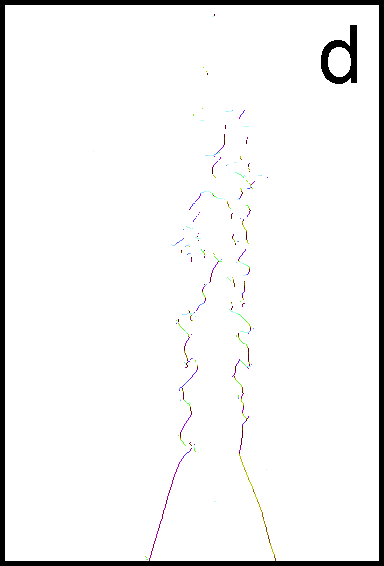}&
		\includegraphics[width=0.180\linewidth]{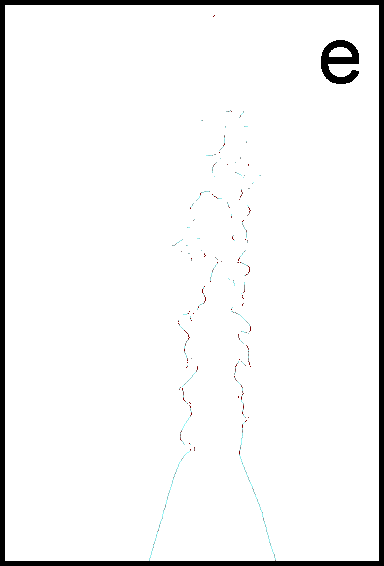}\\	\includegraphics[width=0.180\linewidth]{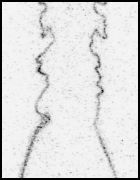}&
		\includegraphics[width=0.180\linewidth]{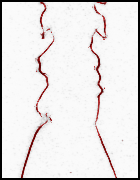}&
		\includegraphics[width=0.180\linewidth]{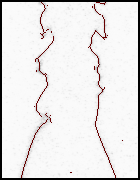}&
		\includegraphics[width=0.180\linewidth]{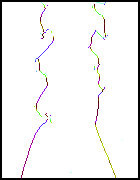}&
		\includegraphics[width=0.180\linewidth]{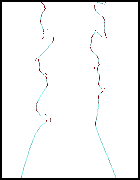}
	\end{tabular}
	\caption{Detection and analysis of flame fronts with CoShREM. (a) PLIF visualization of short-lived {CH} radicals. (b) Pixels colored in dark red correspond to a high value of CoShREM. For illustrative purposes, a brightened version of the processed image is shown in the background. (c) The red lines are obtained from thresholding and thinning the output of CoShREM, depicted in the previous image. (d) Color-coded estimates of the local tangent orientation, where light blue indicates a perfectly horizontal and dark red represents a perfectly vertical orientation. (e) Color-coded estimates of the local curvature, where light blue denotes zero curvature and dark red indicates a curvature greater or equal than $15^{\circ}$. The second row depicts enlarged sections of the images shown in the top row.}
	\label{fig:real_lines}
\end{figure*}

We conclude Section~\ref{sec:results} with two real-world applications of CoShREM. For this purpose, we reuse images recorded in simultaneous single-shot {CH/OH  PLIF} experiments of a turbulent jet flame. A description of the experimental conditions as well as an analysis of the images can be found in a previous article \citep{kie2008}, where the burner and the diagnostic setup as well as the two images are discussed from a combustion point of view in detail. In Figure~\ref{fig:real_edges}, detected flame front locations, local tangent orientations, and local curvature in a PLIF recording of long-lived {OH} radicals are shown. Figure \ref{fig:real_lines} depicts detected flame front locations, local tangent orientations, and local curvature in a PLIF recording of short-lived {CH} radicals. It can be seen that the flame front is picked up reliably in both cases. 

\setlength{\tabcolsep}{2pt}

\renewcommand{\arraystretch}{1}

\section{Conclusion and Outlook}

A novel edge and ridge detection technique for the automated extraction of flame fronts in recordings obtained from imaging techniques such as PLIF or LRS was introduced. Both in the edge and the ridge detection case this method is based on complex-valued anisotropic analyzing elements -- so-called complex shearlets -- and exploits the special scale-independent behavior of the real and imaginary parts of the coefficients associated with these elements at the locations of edges and ridges. Furthermore, the newly proposed CoShREM yields estimates of the local tangent orientations as well as the local curvatures at the locations of edges and ridges.  This new method was also honestly compared to the state of the art.  That is, the authors attempted to optimize the success of the competing methods which were tested.

\

Detection results on PLIF recordings of {OH} and {CH} radicals suggest the applicability of the measures proposed in this paper for real-world data. In addition, numerical experiments on mock images distorted by Gaussian blur and additive Gaussian white noise or Gaussian white noise followed by Poisson noise indicate that the performance of CoShREM is at least visually similar with already established methods such as the properly parameterized Canny edge detector or the phase congruency measure. However, it should be highlighted at this point that for certain applications requiring edge but not ridge detection, all needs might be satisfied by using the well known Canny edge detector or the shearlet-based edge detector more recently proposed by Yi et al. While the former, when given the correct parameters, seems to provide very stable and precise detection results in the presence of noise, the latter also yields approximations of the local tangent orientation and curvature, although not as fine-tuned as our method.

\

The arguably most important feature of CoShREM is that it offers a unified approach to edge and ridge detection while naturally yielding approximations of the local tangent orientations, which can then be used for an effortless computation of local curvature estimates. It hence provides a self-contained package capable of computing many things that might be of interest in the computer-assisted evaluation of experimental data from planar combustion diagnostics. While the current implementation of CoShREM is at least partially outperforming many well-established methods, there is room for future improvement given the relative novelty of the approach.

\

As it is the case with many edge detectors, CoShREM sometimes has difficulties precisely locating corners and intersections, especially when the parameters are chosen in the expectancy of severe distortions. Concerning approximations of the local curvature, there seems to be an imbalance between the curvature detected in the vicinity of corners formed by one vertical and one horizontal edge and the curvature detected around meeting points of two diagonal edges (see e.g. Figure~\ref{fig:mock_edges_orientation_curvature_comparison}). Furthermore, the curvature patterns in Figure~\ref{fig:mock_edges_orientation_curvature_comparison} and Figure~\ref{fig:mock_lines_orientation_curvature_comparison} both show small oscillations -- especially on the detected circles -- that are clearly not present in the analyzed image. However, it seems to be likely that both these issues can be overcome by refining the approximations of the local tangent orientations.

\

The CoShREM-based analysis of the PLIF recording of long-lived OH radicals depicted in Figure~\ref{fig:real_edges} took roughly $9.5$ seconds on a 3.60 GHz Intel Core i7-4790 CPU, where $6.5$ seconds were required for constructing a set of analyzing complex-valued shearlets that can be stored and reused for processing images of the same size. While this seems to be fast enough to put the algorithm to use in most practical situations, it should be noted that processing one image of similar size with the Canny edge detector, Yi et al.'s shearlet-based edge detector or the phase congruency measure only took about 1 second on the same machine. 

\

The current implementation of CoShREM requires the user to set a total of ten different parameters, where six parameters are needed to define a system of complex-valued shearlets and four parameters  configure the actual edge or ridge detection. Especially in comparison to the Canny edge detector and the shearlet-based edge detector introduced by Yi et al., which do not require more than three parameters in the implementations used for this work, this is a lot. However, all of the parameters used in the paper for all of the methods (not just CoShREM) as well as a discussion on how to choose the proper parameters for CoShREM may be found on \url{http://www.math.uni-bremen.de/cda}. Furthermore, we would also like to emphasize again that when testing our method on the mock data, we used the same parameters across all noise levels.  In particular, the parameters remained the same whether or not the mock image was corrupted by Poisson noise.  Also, since CoShREM is implemented with a GUI, it is very easy to modify the parameters. Finally, it seems reasonable that all ten parameters currently required might be derived from a set of only two to three parameters, specifying the expected degree of distortions, the scale of the structures that are to be detected and the expected smoothness of their boundary curves.   

\

Besides the possible improvements outlined in the preceding paragraphs, there is a straightforward generalization of CoShREM to 3D via the three-dimensional shearlet transform \citep{KuLeLi2012,KuLiRei2016,NeLa2012}. While computationally demanding, such a generalization could again significantly improve the detection of flame fronts in high-frequency videos or 3D snapshots of combustion processes. 

\

A Matlab toolbox implementing CoShREM -- including a GUI -- as well as scripts reproducing all results shown in the figures and tables of Section~\ref{sec:results} can be downloaded from \url{http://www.math.uni-bremen.de/cda}.  A video of detected flame fronts is also hosted on the site.

\

\bibliographystyle{spbasic}
\bibliography{shearedge}{}

\end{document}